\documentclass{article}

\PassOptionsToPackage{table}{xcolor}

\usepackage{iclr2027_conference}
\usepackage{times}

\usepackage[T1]{fontenc}
\usepackage{inconsolata}

\usepackage{amsmath,amsfonts,bm}

\def\eqref#1{equation~\ref{#1}}
\def\1{\bm{1}}

\DeclareMathAlphabet{\mathsfit}{\encodingdefault}{\sfdefault}{m}{sl}
\SetMathAlphabet{\mathsfit}{bold}{\encodingdefault}{\sfdefault}{bx}{n}

\usepackage{xcolor}
\usepackage{graphicx}
\usepackage{booktabs}
\usepackage{multirow}
\usepackage{adjustbox}
\usepackage{array}
\usepackage{tabularx}
\usepackage{makecell}
\usepackage{ragged2e}
\usepackage{wrapfig}

\usepackage{algorithm}
\usepackage{algpseudocode}
\usepackage{caption}

\usepackage{enumitem}
\usepackage{listings}
\usepackage[most]{tcolorbox}
\tcbuselibrary{skins,breakable,listings}

\usepackage{titletoc}

\usepackage{hyperref}

\definecolor{figactplan}{HTML}{F4F7FD}
\definecolor{figactground}{HTML}{FFFAF8}
\definecolor{figactact}{HTML}{FBFCF3}
\definecolor{figactink}{HTML}{283241}

\definecolor{figactplandark}{HTML}{667AA8}
\definecolor{figactgrounddark}{HTML}{B98072}
\definecolor{figactactdark}{HTML}{87905C}

\definecolor{FigActDark}{HTML}{536AA3}
\definecolor{FigActLight}{HTML}{F3F5FA}
\definecolor{FigActBorder}{HTML}{A0A9BF}

\definecolor{heatgreen}{HTML}{B9EAB9}

\newcommand{\chg}[2]{%
    #1\,{\scriptsize\textcolor{green!45!black}{#2}}%
}

\newcommand{\chgneg}[2]{%
    #1\,{\scriptsize\textcolor{gray}{#2}}%
}

\newcommand{\hzero}[1]{#1}
\newcommand{\hA}[1]{\cellcolor{heatgreen!12}#1} % 1--5
\newcommand{\hB}[1]{\cellcolor{heatgreen!25}#1} % 6--10
\newcommand{\hC}[1]{\cellcolor{heatgreen!38}#1} % 11--15
\newcommand{\hD}[1]{\cellcolor{heatgreen!52}#1} % 16--20
\newcommand{\hE}[1]{\cellcolor{heatgreen!68}#1} % 21--25
\newcommand{\hF}[1]{\cellcolor{heatgreen!85}#1} % 26--30

\lstdefinestyle{figactjson}{
    basicstyle=\ttfamily\small,
    columns=fullflexible,
    keepspaces=true,
    showstringspaces=false,
    breaklines=true,
    frame=none,
    aboveskip=2pt,
    belowskip=0pt,
    xleftmargin=0pt
}

\title{
    FigAct: Turning Scientific Figures into Active Canvases for Explanation
}

\author{
    \makebox[\dimexpr\textwidth-2\tabcolsep-3pt\relax][c]{%
        \begin{tabular}{c}
            Shishi Xiao\textsuperscript{1}\quad
            Zichao Wang\textsuperscript{2}\quad
            Alexa Siu\textsuperscript{2}\quad
            David H. Laidlaw\textsuperscript{1}\quad
            Jennifer Healey\textsuperscript{2}
            \\[3pt]
            \normalfont\textsuperscript{1}Brown University
            \qquad
            \textsuperscript{2}Adobe Research
        \end{tabular}%
    }
}
\iclrfinalcopy

\begin{document}

\maketitle

% ICLR 的 \maketitle 会写入 “Published as ...” 页眉，
% 因此必须在 \maketitle 之后清空。
\fancyhead{}
\renewcommand{\headrulewidth}{0pt}

\begin{abstract}
Scientific figures are designed to communicate information visually, yet
MLLMs typically explain them by translating their visual content back into
text. This requires readers to manually map the resulting explanations back
to the figure. Inspired by how people present visual information, we introduce
FigAct, a framework that transforms static scientific figures into
question-conditioned visual presentations by acting directly on their
existing graphical elements. Like a human presenter, FigAct generates a
sequence of short narrations, grounds each narration in the corresponding
visual evidence, and applies visual actions to guide the viewer's attention.
We develop a hierarchical search strategy for efficient element localization,
reducing token usage by approximately 40$\times$. We further train FigAct-8B
using three task-specific rewards for grounding accuracy, search efficiency,
and rendering quality. We further build a human-verified benchmark from
figures in real-world scientific papers to evaluate the ability of MLLMs to
generate grounded visual explanations. Our results demonstrate the
effectiveness of FigAct and show that treating scientific figures as
presentation canvases makes explanations clearer and easier to follow.
\end{abstract}

% ============================================================
% Main paper
% ============================================================

\section{Introduction}

% MLLMs can understand and reason about scientific figures → but can they visually communicate that understanding through the figures themselves?
%Scientific figures are designed to communicate visually, yet multimodal large language models still explain them primarily through text~\cite{lu2024mathvista,yue2024mmmu,wang2024charxiv,masry2025chartqapro,pramanick2024spiqa}. Readers are required to map these explanations back to the figure, such as locating the corresponding visual evidence and reconstructing the relations described in the text response. This creates a mismatch between how scientific information is visually organized and how model understanding is communicated.
%Human presentations offer a natural alternative.
% Presenters coordinate verbal explanations with visual evidence, progressively directing attention to relevant content rather than exposing all information at once~\cite{mayer2002multimedia,mayer1999based}.

%presenters progressively direct attention to relevant visual content while introducing corresponding explanations~\cite{mayer2002multimedia,mayer1999based}

\begin{figure}[h]
    \centering
    \includegraphics[width=\textwidth]{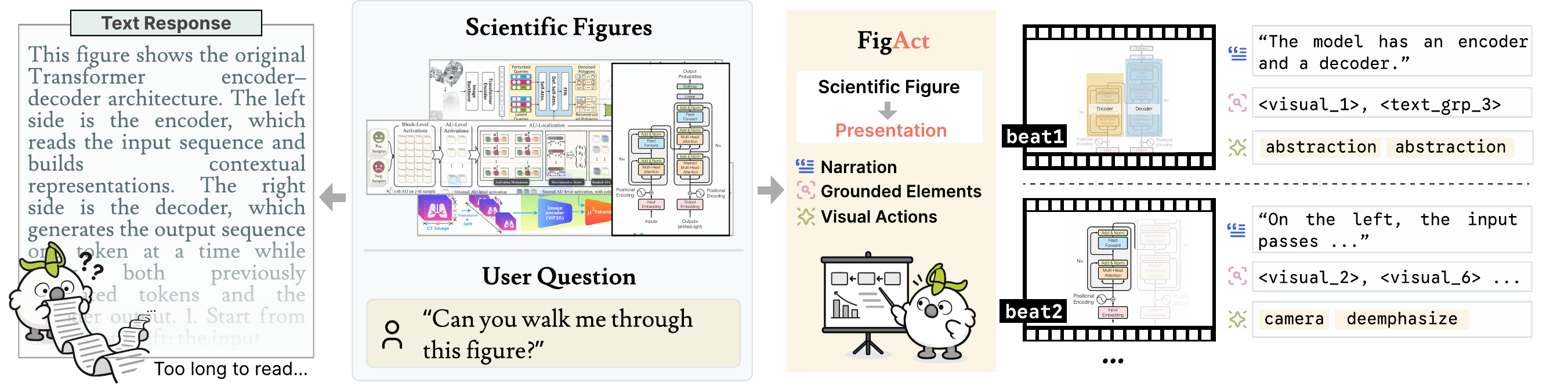}
    \caption{%\textbf{Turning scientific figures into question-conditioned presentation canvases.}
%Instead of returning a textual explanation detached from the figure, FigAct communicates model understanding through the figure itself. Conditioned on a user question, it generates a sequence of presentation beats that align narration, grounded visual evidence, and visual guidance over time.
In response to a user question about a figure, a typical LLM will generate a long text response that is difficult to read and register with the figure.  In contrast, FigAct generates a visual presentation that combines text narration, grounded visual elements, and a sequence of visual actions to guide the viewer through the explanation. See Figure~\ref{fig:result_main} for a concrete example.
}
    \label{fig:teaser}
    \vspace{-1em}
\end{figure}

Scientific figures are often dense and hard to interpret on their own, yet they contain much of the information needed to understand the core concepts of the paper. Currently, multimodal large language models (MLLMs) typically explain figures using text-based answers~\cite{lu2024mathvista,yue2024mmmu,wang2024charxiv,masry2025chartqapro,pramanick2024spiqa} which readers are required to visually map back to the figure. In an interactive setting, like a poster session, an author might use a system diagram to explain their paper to an audience by drawing their attention to different parts of the diagram and explaining the flow of information with a spoken answer and hand gestures~\cite{mayer2002multimedia,mayer1999based}. 
Inspired by this presentation process, we introduce FigAct, which turns a static scientific figure into an interactive visual presentation that responds to user questions. 
As shown in Figure~\ref{fig:teaser}, rather than producing a long text response, FigAct generates an answer as a sequence of visual ``beats.'' Each beat pairs a short narration with the visual evidence it refers to and applies actions such as \texttt{camera} or \texttt{deemphasize} to direct attention. Through a series of these visual beats, FigAct answers the user query by progressively guiding the reader through the figure, similar to how a person might explain it.

We believe that the potential for scientific figures to be used as a medium of communication and storytelling is under-explored. Authors put significant effort into thinking about which elements to include and how to arrange them to convey complex concepts as clearly as possible. In this paper, we investigate how a structured framework can enable MLLMs to transform static figures into a visual story that people can easily follow.
Our proposed framework, FigAct, consists of three stages: Plan, Ground, and Act to model \emph{what should be communicated}, \emph{what should be shown}, and \emph{how it should be presented}. 
\begin{itemize}
    \item \textbf{Plan} organizes the intended text explanation into a sequence of presentation steps,
    \item \textbf{Ground} identifies the relevant visual evidence in the original figure for each step and 
    \item \textbf{Act} determines how this evidence should be transformed to create an optimal guided presentation using a set of visual actions.
\end{itemize}
We use SVG as the working representation of each figure, which provides direct access to individual visual elements. Real scientific SVGs, however, often contain thousands of low-level elements, while the hierarchy produced by authoring tools or format conversion rarely reflects the meaningful visual grouping.
We therefore reorganize them into a geometry-based forest and perform a multi-turn coarse-to-fine search to locate the evidence needed for each beat. Given the grounded evidence, Act selects from a discrete space of high-level presentation actions. A renderer then maps these actions to the grounded elements, separating presentation decisions from low-level visual execution.

In addition to the proposed framework, we train FigAct-8B based on Qwen3-VL-8B~\cite{qwen3technicalreport}. However, obtaining high-quality supervision from real-world scientific figures is difficult because they are visually dense and often lack explicit semantic structure. To train FigAct-8B at scale, we build on VFIG~\cite{he2026vfig}, whose semantically meaningful structure allows us to derive high-quality supervision and convert it into the same trajectories used for real-world SVGs. This yields 32,319 question-presentation pairs from 12,509 figures spanning six question types. We further optimize FigAct-8B with three task-specific rewards for grounding correctness, search efficiency, and rendering quality, while adopting a stage-aware GRPO.

Experiments show that FigAct provides consistent gains as a plug-and-play framework for both open-weight and proprietary MLLMs, improving grounding, action generation, and end-to-end presentation quality while reducing SVG token usage by approximately $40$--$50\times$. After supervised finetuning and reinforcement learning, FigAct-8B maintains a clear advantage over the base model across increasing SVG complexity, search depth, and beat length. The trained model also transfers well from our constructed training data to real-world scientific figures. Human evaluation further confirms the benefits of grounded highlighting and visual actions, showing that the figure itself can become part of the explanation.
\section{Related Work}
\label{sec: related_work}
% scientific figure answering to understanding

%\textbf{From Scientific Figure Understanding to Visual Communication.}
%Multimodal scientific understanding has progressed from visual question answering toward reasoning over increasingly diverse and realistic scientific visuals. Recent benchmarks evaluate reasoning over knowledge-intensive content such as charts, diagrams, and scientific illustrations~\cite{lu2024mathvista,yue2024mmmu,wang2024charxiv,masry2025chartqapro, kembhavi2016diagram}, while Multimodal ArXiv~\cite{li2024multimodal} and SPIQA~\cite{pramanick2024spiqa} further situate such reasoning within scientific papers and their figures. Despite this progress, these tasks predominantly communicate visual understanding through textual outputs. In parallel, recent work has begun transforming document-level scientific content into visual communication artifacts, including slide decks, posters, and presentation videos~\cite{zheng2025pptagent,pang2026paper2poster,shi2025presentagent,liu2025preacher,zhu2025paper2video}. In these systems, scientific figures are treated largely as static content.
%This misses how a human presenter explains a complex figure step by step with dynamic visual guidance.

The key contributions of FigAct are in the combination of its three stages: Plan, Ground and Act. We review prior work relevant to each stage here. Prior work similar to the complex figure reasoning in our Plan stage has been published with respect to reasoning about mathematical figures ~\cite{lu2024mathvista}, charts~\cite{wang2024charxiv,masry2025chartqapro}, diagrams and combinations of these ~\cite{yue2024mmmu, li2024multimodal}. We believe the closest prior work to ours is SPIQA~\cite{pramanick2024spiqa} which focuses specifically on scientific figures in research papers, however, unlike FigAct, all of these works generate text based answers.  Prior work related to the figure transformation in our Act stage can be found in the literature on transforming papers, inclusive of figures, into visual artifacts such as slide decks, posters, presentation videos, and interactive agents~\cite{zheng2025pptagent,pang2026paper2poster,shi2025presentagent,liu2025preacher,zhu2025paper2video, miao2025paper2agent}, however, these transformations largely treat figures as static content or generate new figures from scratch~\cite{ge2025autopresent}.
By contrast, FigAct turns the original figure into an active canvas that is progressively augmented across presentation beats to communicate the answer.

%In these systems, scientific figures are treated largely as static content.
%This misses how a human presenter explains a complex figure step by step with dynamic visual guidance.
% For information-dense scientific figures, treating the figure as static content leaves much of the author's carefully designed visual structure underused for explanation.
%We instead treats the scientific figure itself as an active canvas for explanation, building on the author's original visual design to guide the reader progressively through the answer.
% This misses the step-by-step visual guidance of human presentation. FigAct instead treats the scientific figure as an active canvas for explanation, using its original visual design to progressively guide the reader through the answer.
% Benchmarks such as MathVista~\cite{lu2024mathvista}, MMMU~\cite{yue2024mmmu}, CharXiv~\cite{wang2024charxiv}, and ChartQAPro~\cite{masry2025chartqapro} span charts, diagrams, and other scientific visualizations, while Multimodal ArXiv~\cite{li2024multimodalarxiv} and SPIQA~\cite{pramanick2024spiqa} further link figures with captions and surrounding paper content. Across this line, however, the figure is treated as something to be queried, and both the reasoning target and the output remain predominantly textual.

Both Act and Ground additionally rely on intermediate representations of the figure for fine-grained visual reasoning~\cite{wang2024visually,sun2026geotikzbridge} and temporal explanations~\cite{ku2025theoremexplainagent}. Recent works have explored representing text~\cite{wei2025words}, graphical elements, and semantic relationships~\cite{belouadi2024automatikz,belouadi2024detikzify,yang2026omnidiagram,li2025opusanimation} in scientific figures using structured visual languages such as SVG and TikZ. This has been used for reconstructing existing figures into structured representations and creating editable structures~\cite{lin2026autofigure, liu2026figures, zhao2026crafter, he2026vfig, zeng2026davinci}. 
Inspired by these works, FigAct builds on SVG representation. Their element-level addressability supports fine-grained grounding, while their editability allows Act to manipulate the grounded evidence.

%\textbf{Structured Visual Representations for Scientific Content}
%Structured visual languages represent graphical elements and relations through explicit and executable specifications, enabling precise control over visual structure beyond implicit pixel-space generation~\cite{belouadi2024automatikz,belouadi2024detikzify,wei2025words,ge2025autopresent,yang2026omnidiagram}. Such representations have increasingly been used for complex scientific graphics, from reconstructing existing figures into structured SVG or TikZ~\cite{lin2026autofigure, liu2026figures} to generating publication-ready methodology and pipeline figures with editable structure~\cite{zhao2026crafte, he2026vfig, zeng2026davinci}.
%Beyond visual production, structured graphics can also support multimodal understanding and communication, serving as intermediate representations for fine-grained visual reasoning~\cite{wang2024visually,sun2026geotikzbridge} and temporal explanations~\cite{ku2025theoremexplainagent}.
\section{Problem Formulation}

We study \textbf{Visual Question Presentation (VQP)}, the task of answering a question about a scientific figure through a reader-facing visual presentation.
Given a figure $\mathcal{F}$, consisting of addressable visual elements $\mathcal{E}(\mathcal{F})$, and a user question $q$, the goal is to produce a presentation $\mathcal{P}$ that communicates the answer through narration, grounded visual evidence, and visual actions.

We organize $\mathcal{P}$ as a sequence of \emph{visual beats}, $\mathcal{P}=(b_t)_{t=1}^{T}$.
We borrow the notion of a \emph{beat} from film theory, where it is a minimal unit organized around a coherent dramatic purpose~\cite{el2004interactive}, and use it here to denote one explanatory step in the presentation.
Each beat is given by $b_t=(n_t,G_t,A_t)$, where $n_t$ is a short narration, $G_t\subseteq\mathcal{E}(\mathcal{F})$ is the visual evidence supporting that narration, and $A_t$ defines a set of visual actions applied to the grounded elements.
\textsc{Plan} generates the narration sequence $n_{1:T}$, \textsc{Ground} identifies the corresponding evidence $G_{1:T}$, and \textsc{Act} selects the visual actions $A_{1:T}$ conditioned on the \textsc{Plan} and \textsc{Ground}. This stage-wise generation process is
\begin{equation}
p_{\theta}(\mathcal{P}\mid\mathcal{F},q)
=
\underbrace{p_{\theta}(n_{1:T}\mid\mathcal{F},q)}_{\textsc{Plan}}
\;
\underbrace{p_{\theta}(G_{1:T}\mid n_{1:T},\mathcal{F},q)}_{\textsc{Ground}}
\;
\underbrace{p_{\theta}(A_{1:T}\mid G_{1:T},n_{1:T},\mathcal{F},q)}_{\textsc{Act}}.
\end{equation}

\section{Plan--Ground--Act Framework} \label{sec:framework} 

\begin{figure}[t]
    \centering
    \includegraphics[width=\textwidth]{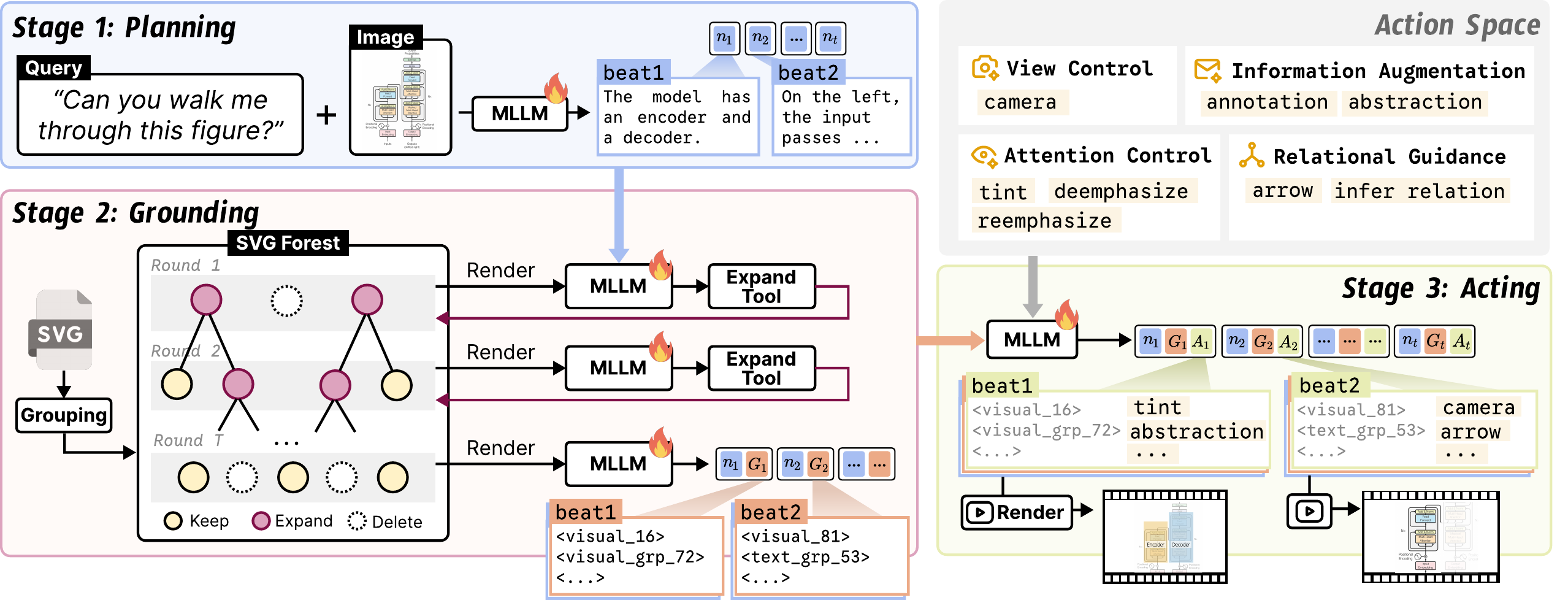}
    \caption{\textbf{Overview of FigAct.} FigAct generates a presentation through three stages. Planning generates the text explanation into a sequence of beats. Grounding groups the SVG into a hierarchical forest and performs multi-round search with \texttt{Keep}, \texttt{Expand}, and \texttt{Delete} operations to identify the visual evidence for each beat. Acting designs presentation actions to the grounded elements and forms the resulting temporal visual presentation.}
    \label{fig:framework_overview}
    \vspace{-1em}
\end{figure}

A human presenter typically explains a figure while directing the audience's attention to the parts that matter at each moment. FigAct models this process with three connected stages as shown in Figure~\ref{fig:framework_overview}. \textsc{Plan} decides what to communicate and in what order, \textsc{Ground} locates the visual evidence for each part of the explanation, and \textsc{Act} determines how that evidence should be presented. 
% As illustrated in Fig.~\ref{fig}, given a figure--question pair $(\mathcal{F},q)$, \textsc{Plan} produces a sequence of narrations $(n_1,\ldots,n_T)$, \textsc{Ground} identifies the supporting SVG elements $G_t$ for each narration $n_t$, and \textsc{Act} selects the corresponding visual actions $A_t$ applied to the grounded evidence. We describe each stage below.

\subsection{Planning: What Should Be Communicated?}
Given the rendered figure $\mathcal{F}$ and question $q$, \textsc{Plan} decomposes the answer into a temporally ordered sequence of short narrations $(n_1,\ldots,n_T)$. Each narration defines one visual beat and determines what should be communicated before the presentation moves to the next step.

\subsection{Grounding: What Should Be Shown?}
Given a narration $n_t$, \textsc{Ground} identifies the SVG elements $G_t \subseteq \mathcal{E}(\mathcal{F})$ that provide the visual evidence for that beat. SVG is particularly useful for this purpose because its elements are individually addressable, any subset can be rendered for visual inspection, and the selected elements remain editable for subsequent presentation actions. The main challenge is scale. Real scientific figures often contain hundreds or thousands of low-level SVG primitives, while their native structure provides little guidance about which elements form meaningful visual units.

\noindent \textbf{Building the Hierarchical Search Space.} 
In our real-world corpus, each SVG contains on average 302 non-text \texttt{<path>} elements and 26 embedded raster images. Semantically meaningful objects such as arrows, boxes, and text regions may be fragmented across generic SVG primitives, and the hierarchy produced by authoring or format conversion rarely follows the visual organization perceived by a reader. We therefore reorganize the SVG into a geometry-based forest.
We first aggregate low-level primitives into local visual groups. Text elements are clustered by spatial proximity using DBSCAN~\cite{ester1996density}, while non-text primitives are grouped using their style and geometric properties. We then establish parent--child relations through spatial containment, assigning each group to its smallest enclosing parent. The resulting forest provides a coarse-to-fine view of the figure, from larger visual regions to progressively finer groups and individual SVG elements.

\noindent
\begin{minipage}[t]{0.70\linewidth}
\vspace{0pt}
\noindent\textbf{Rendering Candidates for Visual Inspection.}
At each level of the forest, the current nodes form the candidate set inspected by the MLLM. We render these candidates in two complementary views. For the context view, we follow Set-of-Mark prompting~\cite{yang2023setofmark} and overlay candidate identifiers on the full figure, preserving the spatial context needed to interpret each candidate. We additionally exploit the renderability of SVG to create an isolated view, where each candidate is rendered separately with the same identifier. This allows the model to inspect its visual content without interference from surrounding elements. The isolated candidates are arranged adaptively according to their rendered size to reduce image-token usage.
\end{minipage}
\hfill
\begin{minipage}[t]{0.28\linewidth}
\vspace{0pt}
\centering
\includegraphics[width=\linewidth]{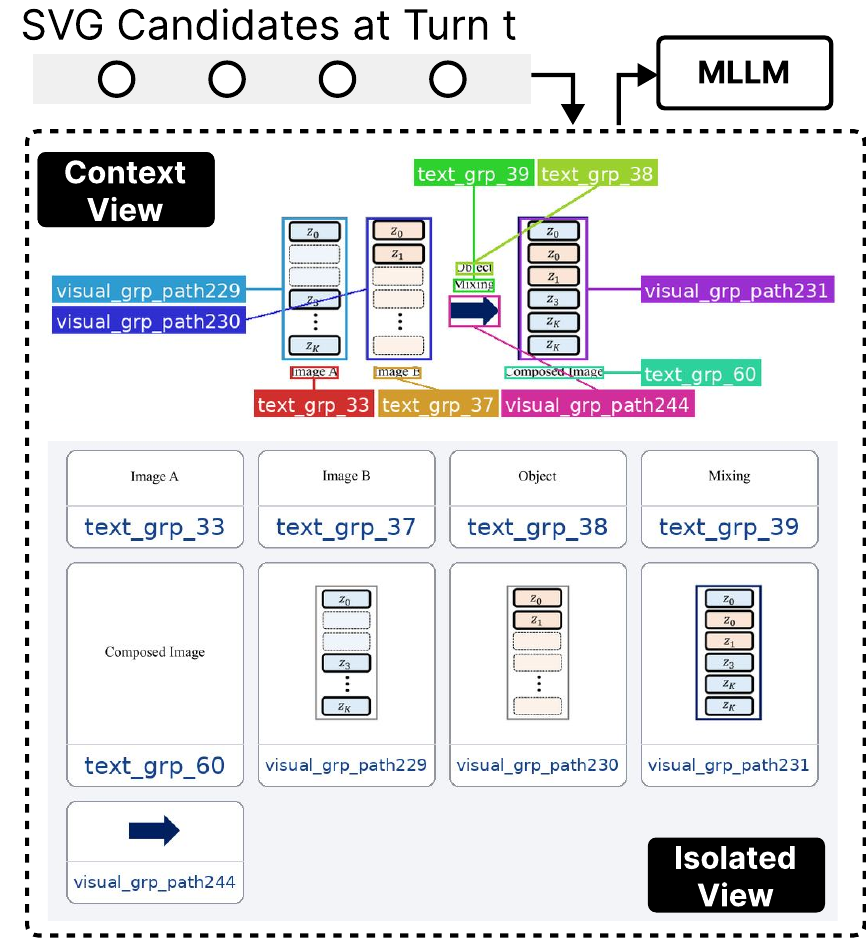}
\label{fig:candidate_rendering}
\end{minipage}

\noindent \textbf{Hierarchical Grounding Trajectory.}
Grounding is performed as a multi-turn search over this forest. At each turn, the MLLM inspects the currently visible candidates and assigns one of three operations to every node: \texttt{Keep}, \texttt{Expand}, or \texttt{Delete}. \texttt{Keep} selects the node and its covered SVG elements as grounding evidence. \texttt{Expand} opens the node for finer inspection by exposing its children in the next turn. \texttt{Delete} removes the branch from further search. Only expanded nodes are revisited, allowing the model to progressively narrow its search while avoiding irrelevant regions of the figure. 
The trajectory ends when no candidate is further expanded, and $G_t$ is given by the SVG elements covered by the retained nodes.

\subsection{Acting: How It Should Be Presented?}
Given the narration $n_t$ and grounded evidence $G_t$, the acting stage determines how the evidence should be visually presented. Prior work~\cite{lin2024showui,lin2026jarvisart,zheng2025tracevla} has shown the value of representing complex behavior with atomic operations and an explicit action space, allowing domain-specific knowledge to be encoded into structured decisions. Following this principle, we derive this action vocabulary from recurring visual strategies observed in publicly available presentation videos. It covers four presentation functions, including \textbf{view control} (\texttt{camera}), \textbf{attention control} (\texttt{tint}, \texttt{deemphasize}, \texttt{reemphasize}), \textbf{relational guidance} (\texttt{arrow}, \texttt{infer\_relation}), and \textbf{information augmentation} (\texttt{annotation}, \texttt{abstraction}).  For each beat, the MLLM produces an ordered action sequence $A_t=(a_{t,1},\ldots,a_{t,K_t})$, where each action specifies an operation, its target SVG elements, and the customized parameters. A deterministic renderer maps these high-level decisions to low-level SVG transformations and executes them to produce the final animation, allowing the MLLM to focus on how the evidence should be presented, while the renderer handles visual execution.
This separation is useful because the model is responsible for the semantic presentation decision, while the renderer takes care of the low-level implementation. 
More details are provided in Appendix~\ref{supp: method_detail}.
\section{Two-Stage Training}

\subsection{Stage 1: SFT Warm-Up}

We first perform structured SFT to initialize FigAct-8B with the complete Plan--Ground--Act workflow. 
% Given a scientific figure and question, the model learns to generate a sequence of narration beats, ground each beat through the hierarchical SVG search process, and produce the corresponding visual actions. 
Outputs from all three stages are represented in a unified JSON-based format, providing a structured initialization for subsequent RL training.

\subsection{Stage 2: RL Training}
\label{sec:rl}
Following SFT, we further use reinforcement learning to optimize FigAct-8B, grounding visual evidence and applying visual actions for these grounded results.
For input including figure $\mathcal{F}$, question $q$, with fixed plan $n_{1:T}$, we sample $K$ Ground--Act trajectories from current policy $\theta_{old}$
\begin{equation}
    \tau_i=(G_{1:T}^{(i)},A_{1:T}^{(i)})
    \sim
    \pi_{\theta_{old}}(\cdot \mid \mathcal{F}, q, n_{1:T}),
    \qquad i=1,\ldots,K ,
\label{eq:rl_rollout}
\end{equation}
where the predicted grounding $G_{1:T}^{(i)}$ conditions the subsequent action generation $A_{1:T}^{(i)}$.
Each trajectory is evaluated along three complementary dimensions: whether the correct evidence is found, how efficiently it is found, and whether the resulting presentation visually communicates the narration.

\textbf{Grounding Correctness Reward.}
As grounded elements are identified at different levels of the SVG forest over multiple rounds of hierarchical search, we canonicalize each grounding result into its terminal element set, recursively replacing expanded elements with their terminal descendants.
For the $i$-th sampled output and $t$-th presentation beat, let $\hat{G}_{i,t}$ and $G_t^*$ denote the predicted and reference terminal element sets, respectively. Correctly grounded elements are those appearing in both sets. We compute element-level precision and recall as
\begin{equation}
    \mathsf{Prec}_{i,t}
    =
    \frac{|\hat{{G}}_{i,t}\cap{G}_t^*|}
         {|\hat{{G}}_{i,t}|},
    \qquad
    \mathsf{Rec}_{i,t}
    =
    \frac{|\hat{{G}}_{i,t}\cap{G}_t^*|}
         {|{G}_t^*|},
    \qquad
    r_i^{\mathrm{grd}}
    =
    \frac{1}{T}
    \sum_{t=1}^{T}
    \frac{2\,\mathsf{Prec}_{i,t}\mathsf{Rec}_{i,t}}
         {\mathsf{Prec}_{i,t}+\mathsf{Rec}_{i,t}} .
\label{eq:ground_reward}
\end{equation}

\textbf{Search Efficiency Reward.}
Two search trajectories may recover the same grounded elements while exploring very different portions of the SVG hierarchy.
We therefore reward Ground for reaching the target evidence with lower search cost.
We define the search cost $C$ as the total number of child nodes revealed by expansion along a trajectory, which measures the intermediate elements exposed during search to reach the final grounded elements.
Let $\hat{C}_i$ and $C_i^*$ denote the predicted and reference search costs, respectively.
We define the search reward as
\begin{equation}
r_i^{\mathrm{search}}
=
\left(r_i^{\mathrm{grd}}\right)^2
\min\left(
1,
\frac{C_i^*+1}{\hat{C}_i+1}
\right).
\label{eq:search_reward}
\end{equation}
The grounding term ensures that a short but inaccurate search does not receive a high reward.

\textbf{Rendering-Aware Reward.}
Following the Act stage, a deterministic renderer converts the predicted high-level visual actions into executable SVG animations, which are rendered into videos. 
We measure the semantic alignment between each rendered beat $V_{i,b}$ and its corresponding narration $n_b$ using the frozen video--text encoder PE-Core~\cite{bolya2026perception}. The rendering-aware reward is averaged across all $B$ presentation beats:
\begin{equation}
    r_i^{\mathrm{render}}
    =
    \frac{1}{T}
    \sum_{t=1}^{T}
    \cos\!\left(
        \phi_V(V_{i,t}),
        \phi_T(n_t)
    \right).
\label{eq:render_reward}
\end{equation}

\textbf{Stage-Aware Reward Decoupling.}
The three rewards capture different properties of a Ground--Act trajectory and may follow different distributions across rollouts. Directly summing them before group normalization can map distinct reward profiles to the same scalar score, weakening the learning signal from each component. Following GDPO~\cite{liu2026gdpogrouprewarddecouplednormalization}, we therefore normalize each reward independently within the rollout group:
\begin{equation}
a_i^k
=
\frac{r_i^k-\mu_k}{\sigma_k+\epsilon},
\qquad
k\in\{\mathrm{grd},\mathrm{search},\mathrm{render}\}.
\label{eq:gdpo_norm}
\end{equation}
We then assign these normalized rewards according to the stage they evaluate. Ground receives the grounding, search, and rendering signals, since its evidence selection affects both search behavior and the final rendered presentation. Act receives the rendering signal, which directly evaluates the visual result of its action decisions:
\begin{equation}
\tilde{A}_i^{G}
=
\lambda_g a_i^{\mathrm{grd}}
+
\lambda_s a_i^{\mathrm{search}}
+
\lambda_r a_i^{\mathrm{render}},
\qquad
\tilde{A}_i^{A}
=
a_i^{\mathrm{render}} .
\label{eq:stage_aggregation}
\end{equation}
Finally, we apply batch-wise normalization separately to the Ground and Act aggregates, producing $A_i^G$ and $A_i^A$, which are assigned to the corresponding stage tokens during policy optimization.

\section{Dataset}
\label{sec: data}

\subsection{Real-World Scientific Data Collection}
\label{sec: data_real}
\begin{figure*}[t]
\centering

% ---------- (a) ----------
\begin{minipage}[t]{0.34\textwidth}
    \vspace{0pt}
    \centering
    \includegraphics[
        width=\linewidth,
        trim=8 5 8 5,
        clip
    ]{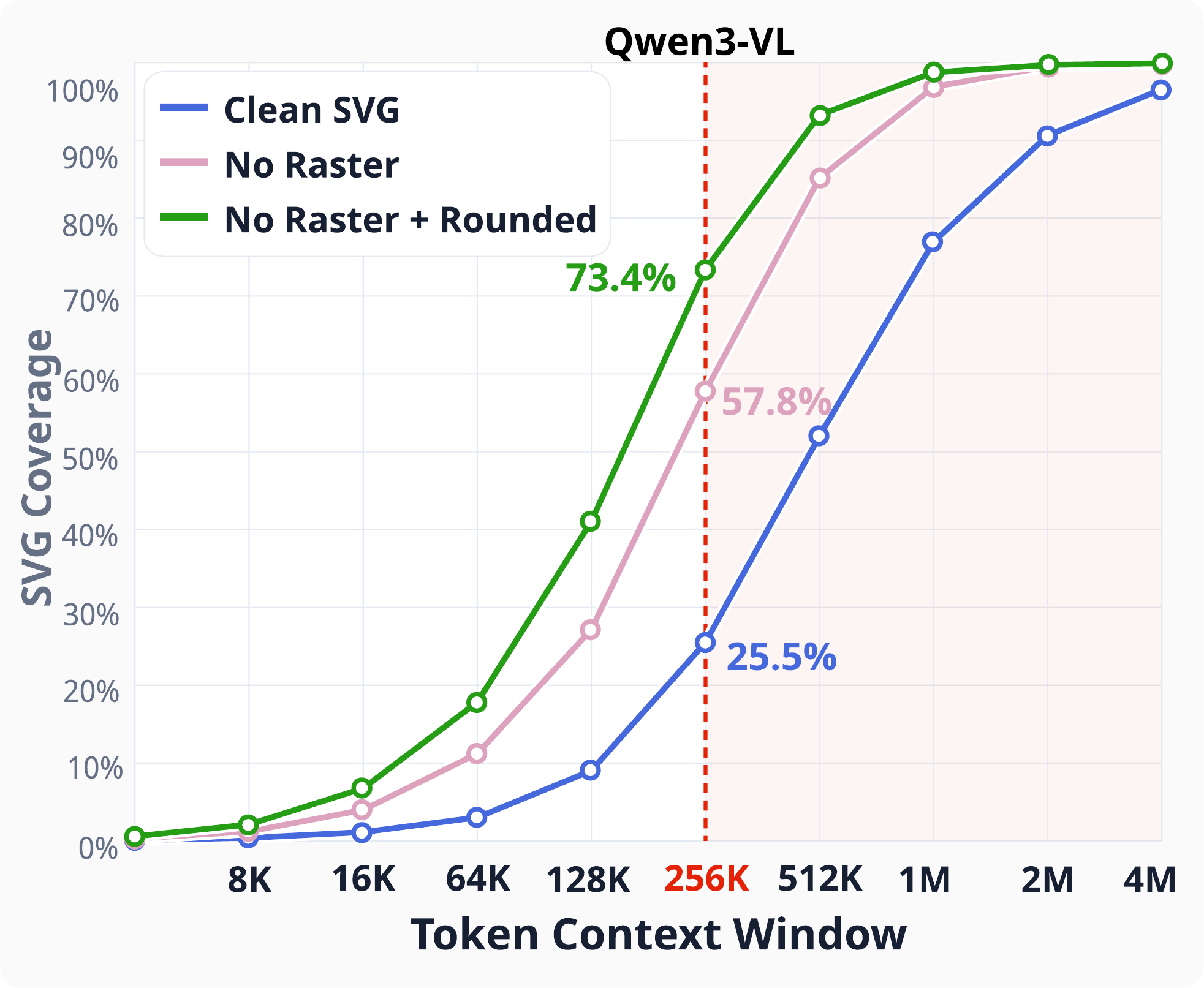}
\end{minipage}%
\hspace{0.01\textwidth}%
%
% ---------- (b) ----------
\begin{minipage}[t]{0.25\textwidth}
    \vspace{0pt}
    \centering
    \includegraphics[
        width=\linewidth,
        trim=12 5 12 5,
        clip
    ]{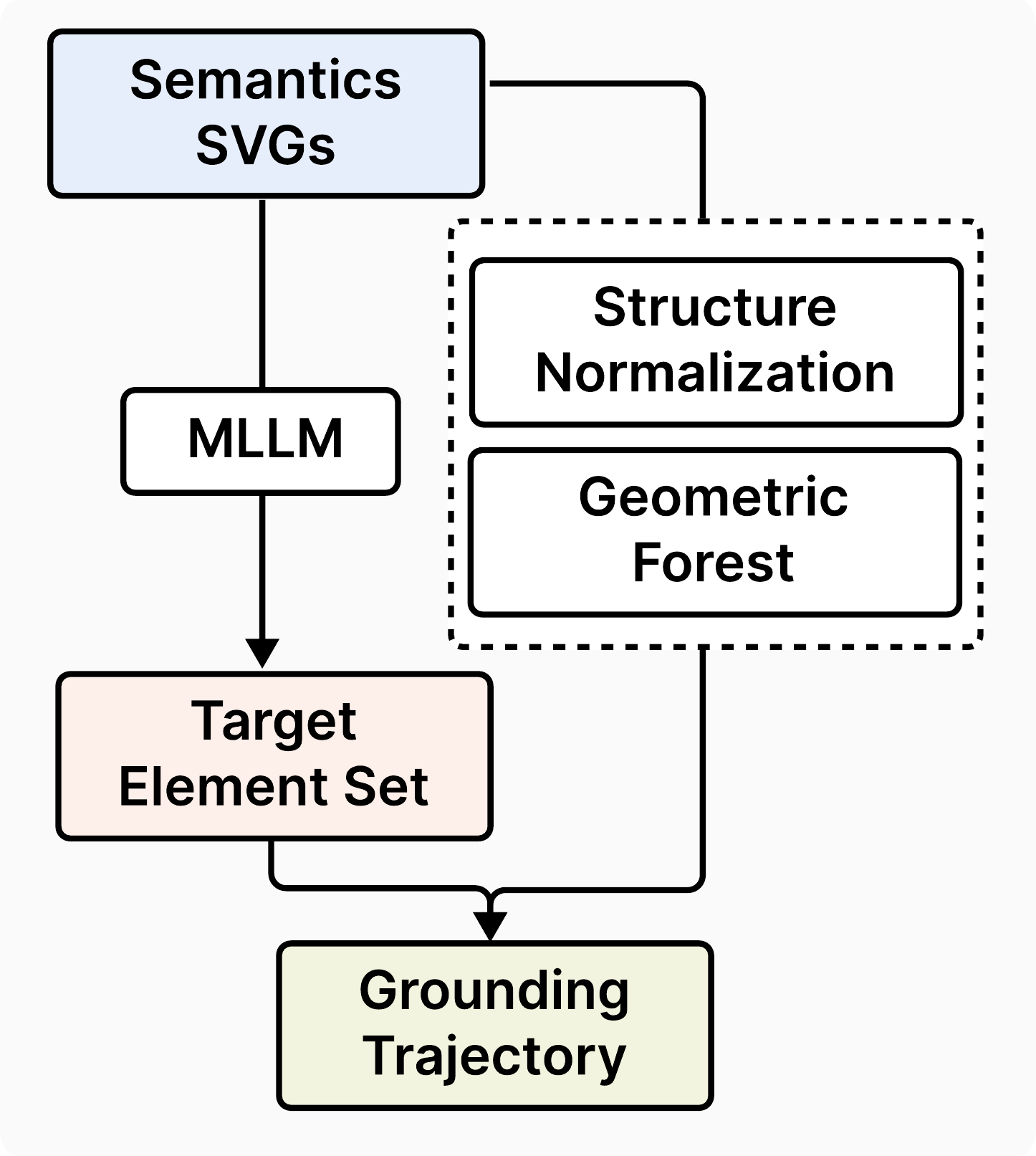}
\end{minipage}%
\hspace{0.01\textwidth}%
%
% ---------- (c) ----------
\begin{minipage}[t]{0.37\textwidth}
    \vspace{0pt}
    \centering

    {\footnotesize
    \setlength{\tabcolsep}{0pt}
    \renewcommand{\arraystretch}{1.22}
    \setlength{\aboverulesep}{0pt}   
    \setlength{\belowrulesep}{0pt}  
    
    \begin{adjustbox}{max width=\linewidth}
    \begin{tabular}{@{}l| c ccc@{}}
    \toprule
    \multirow{2}{*}{\textbf{Task}}
    & \multirow{2}{*}{\textbf{N}}
    & \multicolumn{1}{c}{\textbf{P}}
    & \multicolumn{2}{c}{\textbf{B}} \\
    \cmidrule(lr){3-3}
    \cmidrule(lr){4-5}
    &
    & \textbf{\# Beats}
    & \textbf{\# Elem.}
    & \textbf{\# Act.} \\
    \midrule
    D1 Overview    & 8,419 & 7.8 & 7.6  & 2.9 \\
    D2 Focus       & 4,922 & 3.5 & 10.1 & 4.4 \\
    D3 Rationale   & 4,500 & 3.7 & 10.8 & 4.3 \\
    D4 Relation    & 4,623 & 3.5 & 10.5 & 4.5 \\
    D5 Comparison  & 4,803 & 3.6 & 10.9 & 4.2 \\
    D6 Abstraction & 5,052 & 4.0 & 11.8 & 4.4 \\
    \midrule
    \textbf{Overall}
    & \textbf{32,319}
    & \textbf{4.8}
    & \textbf{9.4}
    & \textbf{3.7} \\
    \bottomrule
    \end{tabular}
    \end{adjustbox}
    }

    \vspace{-3pt}

    {\scriptsize P: presentation; B: beat.}

\end{minipage}
\caption{
\textbf{Synthetic data construction and statistics.}
(a) SVG coverage across context-window sizes under different preprocessing settings.
(b) Construction of grounding trajectories from semantic SVGs.
(c) Statistics of the six task types, reporting sample counts and average presentation- and beat-level quantities.
}
\label{fig:dataset_overview}
\vspace{-1.5em}
\end{figure*}

We collect scientific figures from 2025--2026 papers across major HCI, AI, medical, and robotics venues, restricting our collection to arXiv versions released under the CC BY 4.0 license. We extract figure assets from the \LaTeX{} source and convert PDF figures to SVG, yielding 1,666 real-world scientific SVGs. These SVGs exhibit substantial structural variation due to differences in authoring tools and conversion. Elements are often weakly structured or entirely flat, text is encoded as glyph paths, and raster images are embedded in figures. On average, each figure contains 302 non-text SVG elements, including 26 raster images, together with 461 text glyph paths. As shown in the left panel of Fig.~\ref{fig:dataset_overview}, only 25.5\% of the raw SVGs fit within Qwen3-VL's 256K-token context window. After raster removal and path simplification, this fraction increases to 73.4\%. From this collection, we curate 400 figures for evaluation using an LLM-assisted annotation pipeline. For each figure, GPT-4o generates a user question, Gemini 3.5 Flash and GPT-5.6 produce an initial Plan--Ground--Act annotation, and a human designer reviews and refines both the question and the annotation.

\subsection{Constructing FigAct Supervision}

Obtaining reliable grounding supervision directly from real-world scientific SVGs is difficult because they lack consistent semantic labels and structural hierarchies. We therefore construct training supervision from VFIG-Data-Complex-Diagrams~\cite{he2026vfig}, whose SVGs provide semantic element labels and structured DOM hierarchies. After MLLM-based quality filtering, we retain 12,509 figures.
We construct questions across six representative types---Overview, Focus, Rationale, Relation, Comparison, and Abstraction---covering common ways users may query a scientific figure. As illustrated in  Fig.~\ref{fig:dataset_overview} middle, for each question, we use Qwen3-VL-32B together with the semantic SVG information to generate a beat-level presentation and identify the target visual elements and actions for each beat.
The semantic labels and original DOM hierarchy are used only to construct supervision. We remove both before training, reconstruct each SVG as a geometry-based forest, and project the annotated target elements onto this forest to derive multi-turn \texttt{Keep}, \texttt{Expand}, and \texttt{Delete} grounding trajectories. This yields 32,319 presentation samples with Plan, Ground, and Act supervision across the six question types. Dataset statistics are summarized in the Fig.~\ref{fig:dataset_overview} right.

\section{Experiments}
\label{sec:experiments}

\begin{figure}[t]
    \centering
    \includegraphics[width=\textwidth]{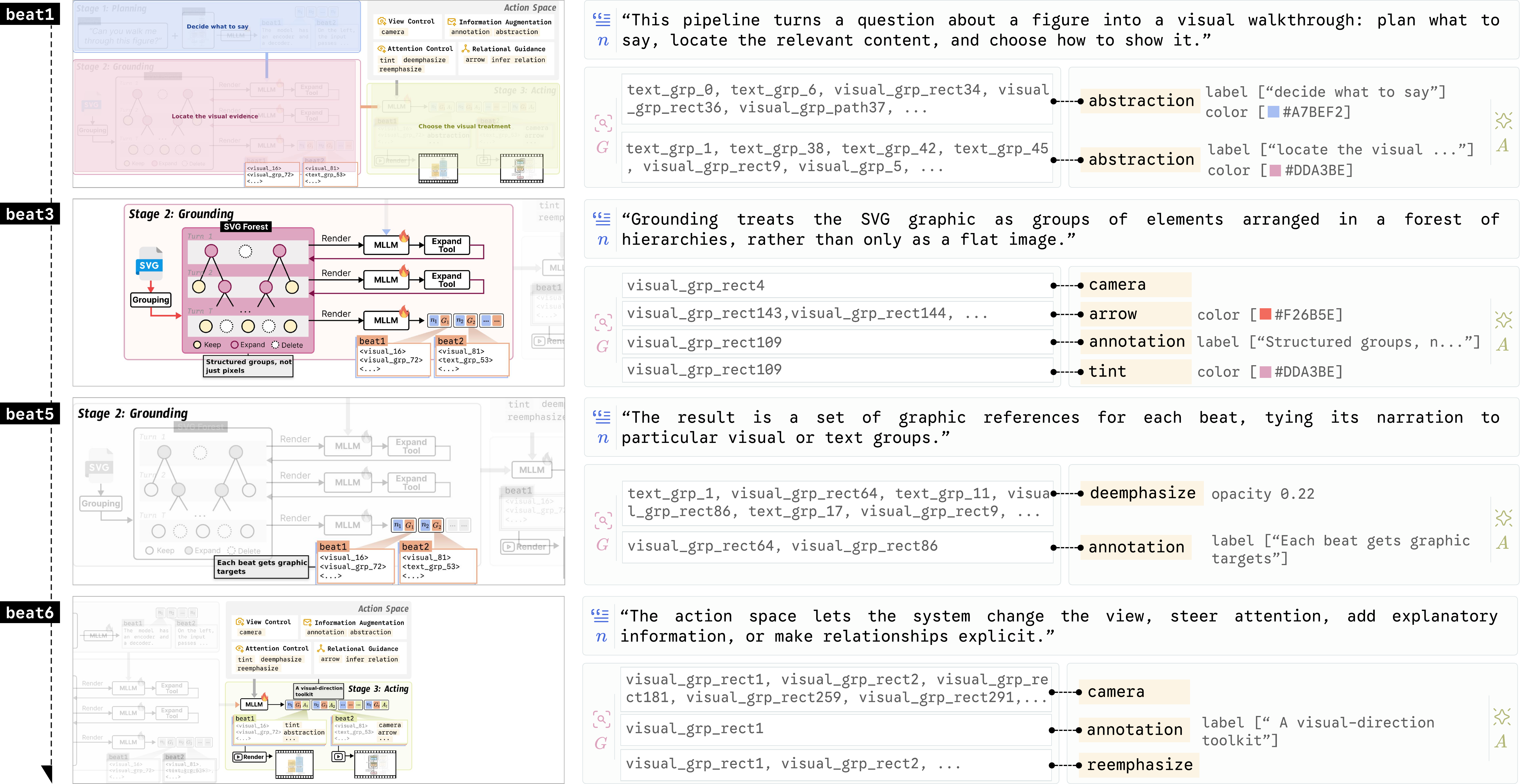}
    \caption{\textbf{Example of Generated Visual Presentation.} We use Figure~\ref{fig:framework_overview} as an example. For the complete presentation, see Appendix~\ref{supp: qualitative}.}
    \label{fig:result_main}
    \vspace{-1em}
\end{figure}

\subsection{Experimental Setup}
\label{sec:exp_setup}

\textbf{Benchmark.}
We evaluate the proposed framework FigAct and trained model FigAct-8B on the real-world benchmark introduced in Sec.~\ref{sec: data_real}, consisting of 400 question--figure pairs with one question per scientific figure.
% The figures are collected from recent papers across multiple scientific domains.

\textbf{Implementation details.}
We initialize FigAct-8B from Qwen3-VL-8B-Instruct and train rank-16 LoRA adapters. SFT is performed for three epochs with a learning rate of $10^{-4}$ and a global batch size of 32.
Starting from the SFT model, we then apply GDPO normalization and stage-specific reward assignment for 500 updates at a learning rate of $10^{-6}$.
Each update uses two presentation instances, with $K=8$ rollout trajectories per instance.
Each RL configuration is trained on eight NVIDIA A100 80\,GB GPUs, with four GPUs used for policy optimization and rollout generation and four for reward computation when required.

\textbf{Baselines.}
We compare against representative proprietary MLLMs, including Claude Fable 5.1~\cite{anthropic2026fable}, Gemini 3.5 Flash~\cite{google2026gemini35flash}, and GPT-6 Astra~\cite{openai2026gpt6astra}, as well as open-weight MLLMs, including Qwen3.5-9B~\cite{qwen3.5}, InternVL3.5-8B~\cite{wang2025internvl3_5}, and Qwen3-VL-8B-Instruct~\cite{qwen3technicalreport}.
For each backbone, we additionally evaluate the performance of integrating FigAct framework.
All methods follow the same multi-turn evaluation protocol with reference propagation.

\textbf{Metrics.}
We evaluate performance at each stage of the Plan--Ground--Act pipeline as well as on the final rendered presentation.
For \textsc{Plan}, we measure the Factuality~\cite{xu-etal-2023-critical} and Coverage~\cite{scialom-etal-2021-questeval} of the generated narration.
For \textsc{Ground}, we report grounding completion rate (Comp.), element-level F1 (Elem. F1), token cost (Tok.), and search efficiency (Search-E).
Search efficiency is reported only for methods that perform hierarchical SVG search.
For \textsc{Act}, we measure execution success rate (Exec.), which evaluates whether the generated actions can be executed over their referenced SVG elements, and Narration--Video Alignment (NVA), computed using Qwen3-VL-Embedding~\cite{qwen3vlembedding}.
Finally, we evaluate the rendered presentation end to end using Presentation Quality (Pres-Q), where a video-LLM judge assesses content quality, visual effectiveness, and temporal coherence.

\begin{table*}[t]
\centering
\small
\setlength{\tabcolsep}{3.6pt}
\renewcommand{\arraystretch}{1.08}
\resizebox{\textwidth}{!}{
\begin{tabular}{l
    cc
    cccc
    cc
    c}
\toprule
& \multicolumn{2}{c}{\textbf{Plan}}
& \multicolumn{4}{c}{\textbf{Ground}}
& \multicolumn{2}{c}{\textbf{Act}}
& \multicolumn{1}{c}{\textbf{End-to-End}} \\
\cmidrule(lr){2-3}
\cmidrule(lr){4-7}
\cmidrule(lr){8-9}
\cmidrule(lr){10-10}

\textbf{Method}
& Fact. $\uparrow$
& Cover. $\uparrow$
& Comp. $\uparrow$
& Elem. F1 $\uparrow$
& Tok. (M) $\downarrow$
& Search-E $\uparrow$
& Exec. $\uparrow$
& NVA $\uparrow$
& Pres-Q $\uparrow$ \\
\midrule

\multicolumn{10}{c}{\textit{Proprietary MLLMs}} \\

\rowcolor{gray!8}
Claude Fable 5.1
& 90.64 & \textbf{90.80}
& 71.00 & 43.44 & 1.12 & --
& 77.06 & 70.88 & 79.52 \\

\quad + FigAct
& -- & --
& \chg{98.00}{+27.00}
& \chg{79.69}{+36.25}
& \chg{0.022}{-1.098}
& 85.96
& \chg{81.32}{+4.26}
& \chg{73.62}{+2.74}
& \chg{84.45}{+4.93} \\

\rowcolor{gray!8}
Gemini 3.5 Flash
& 94.60 & 85.20
& 50.00 & 22.58 & 0.99 & --
& 78.25 & 71.58 & 81.23 \\

\quad + FigAct
& -- & --
& \chg{96.00}{+46.00}
& \chg{68.85}{+46.27}
& \chg{0.015}{-0.975}
& \textbf{95.35}
& \chg{96.71}{+18.46}
& \chg{73.30}{+1.72}
& \chg{\textbf{87.62}}{+6.39} \\

\rowcolor{gray!8}
GPT-6 Astra
& \textbf{99.71} & 90.00
& 81.00 & 49.59 & 1.09 & --
& 82.41 & 72.08 & 85.65 \\

\quad + FigAct
& -- & --
& \chg{\textbf{100.00}}{+19.00}
& \chg{\textbf{82.46}}{+32.87}
& \chg{0.019}{-1.071}
& 74.20
& \chg{\textbf{98.05}}{+15.64}
& \chg{\textbf{73.64}}{+1.56}
& \chg{87.56}{+1.91} \\

\midrule
\multicolumn{10}{c}{\textit{Open-Weight MLLMs}} \\

\rowcolor{gray!8}
Qwen3.5-9B
& 92.12 & 90.73
& 18.00 & 12.03 & 0.16 & --
& 72.26 & 67.23 & 74.66 \\

\quad + FigAct
& -- & --
& \chg{98.00}{+80.00}
& \chg{43.90}{+31.87}
& \chg{\textbf{0.004}}{-0.156}
& 64.45
& \chg{81.93}{+9.67}
& \chg{68.48}{+1.25}
& \chg{81.62}{+6.96} \\

\rowcolor{gray!8}
InternVL3.5-8B
& 93.60 & 89.05
& 0.00 & N/A & N/A & N/A
& 54.18 & 65.94 & 72.65 \\

\quad + FigAct
& -- & --
& \chg{97.00}{+97.00}
& 31.56
& 0.007
& 63.64
& \chg{76.85}{+22.67}
& \chgneg{65.58}{-0.36}
& \chg{81.83}{+9.18} \\

\rowcolor{gray!8}
Qwen3-VL-8B
& 88.69 & 86.32
& 8.00 & 9.85 & 0.16 & --
& 68.63 & 67.92 & 70.18 \\

\quad + FigAct
& -- & --
& \chg{98.00}{+90.00}
& \chg{28.35}{+18.50}
& \chg{\textbf{0.004}}{-0.156}
& 68.26
& \chg{79.13}{+10.50}
& \chg{68.32}{+0.40}
& \chg{78.56}{+8.38} \\

FigAct-8B (SFT)
& -- & --
& \chg{\textbf{100.00}}{+2.00}
& \chg{52.67}{+24.32}
& \textbf{0.004}
& \chg{83.04}{+14.78}
& \chg{95.66}{+16.53}
& \chg{69.68}{+1.36}
& \chg{82.47}{+3.91} \\

\rowcolor{green!10}
FigAct-8B (SFT+RL)
& -- & --
& \textbf{100.00}
& \chg{62.27}{+9.60}
& \textbf{0.004}
& \chg{84.68}{+1.64}
& \chg{97.10}{+1.44}
& \chg{72.25}{+2.57}
& \chgneg{82.16}{-0.31} \\

\bottomrule
\end{tabular}
}
\caption{
\textbf{Main results on real-world scientific figure presentation.}
Across proprietary and open-weight MLLMs, applying FigAct at inference time consistently improves grounding, action generation, and end-to-end presentation quality while substantially reducing SVG token usage. SFT and RL further improve grounding, search efficiency, and action generation for the Qwen3-VL-8B-based FigAct model.
}
% \vspace{-2em}
\label{tab:main_results}
\end{table*}

\subsection{Main Results}
\label{sec:main_results}
We report the main quantitative results in Table~\ref{tab:main_results}, with qualitative presentation examples shown in Fig.~\ref{fig:result_main}. We make three main observations.
(1) \textbf{FigAct provides consistent gains as a plug-and-play framework.} Simply using the FigAct framework at inference time, without additional training, improves \textsc{Ground}, \textsc{Act}, and end-to-end presentation quality across both proprietary and open-weight MLLMs.
Gemini 3.5 Flash, for example, gains $+46.27$ Elem. F1 and $+18.46$ execution success. 
These stage-level gains also translate to the final presentation: Pres-Q improves for every backbone, with gains ranging from $+0.02$ to $+0.09$.
(2) \textbf{Hierarchical search makes fine-grained grounding substantially more token-efficient.} For proprietary MLLMs, FigAct reduces SVG token usage by more than $50\times$ while simultaneously improving grounding completion and Elem. F1. We observe the same pattern for open-weight models, where Qwen3.5-9B and Qwen3-VL-8B each use approximately $40\times$ fewer SVG tokens. This suggests that FigAct does not gain grounding accuracy by exposing the model to more of the SVG; instead, hierarchical search allows it to reach the relevant elements with substantially less context.
(3) \textbf{Training further strengthens FigAct-8B, with SFT contributing the largest gains and RL providing additional improvements.}
On Qwen3-VL-8B, SFT adds $+24.32$ Elem. F1 and $+14.78$ Search-E over the plug-and-play FigAct configuration, while RL further improves Elem. F1 by $+9.60$ and NVA by $+2.57$. The final model reaches $97.10$ execution success and $72.25$ NVA, close to the strongest proprietary FigAct results on both metrics. Pres-Q rises by $3.91$ after SFT and drops by $0.31$ after RL.

\subsection{Ablation Analysis}

\begin{wraptable}{r}{0.45\columnwidth}
    \centering
    \vspace{-8pt}
    \scriptsize
    \setlength{\tabcolsep}{2.6pt}
    \renewcommand{\arraystretch}{1.05}

    \resizebox{\linewidth}{!}{
    \begin{tabular}{lcccc}
        \toprule
        \textbf{Setting}
        & Elem. F1 $\uparrow$
        & Search-E $\uparrow$
        & NVA $\uparrow$
        & Pres-Q $\uparrow$ \\
        \midrule

        SFT only
            & 52.67
            & 83.04
            & 69.68
            & \textbf{82.47} \\

        w/o $R_{\mathrm{ground}}$
            & 60.98
            & 84.15
            & \underline{71.91}
            & 81.97 \\

        w/o $R_{\mathrm{search}}$
            & 60.05
            & 83.28
            & 70.85
            & 82.05 \\

        w/o $R_{\mathrm{render}}$
            & \underline{61.88}
            & \textbf{84.89}
            & 69.35
            & \underline{82.78} \\

        \textbf{Full}
            & \textbf{62.27}
            & \underline{84.68}
            & \textbf{72.25}
            & 82.16 \\

        \bottomrule
    \end{tabular}
    }

    \caption{
        Ablation of RL rewards.
    }
    \label{tab:reward_ablation}
    \vspace{-8pt}
\end{wraptable}

\textbf{Ablation of Rewards}
\label{sec:reward_ablation}
Table~\ref{tab:reward_ablation} shows that reinforcement learning provides clear gains over SFT alone, improving Elem.~F1 from $52.67$ to $62.27$ and also yielding consistent improvements in NVA and Pres-Q. Removing individual rewards reveals complementary effects across stages. Without $R_{\mathrm{search}}$, both grounding accuracy and Search-E drop, suggesting that encouraging efficient exploration also helps the model reach the correct visual evidence. Removing $R_{\mathrm{render}}$ causes the largest degradation in NVA, consistent with its role in providing feedback from the rendered presentation. $R_{\mathrm{ground}}$ further contributes to grounding accuracy and downstream presentation quality. Combining all three rewards gives the strongest overall performance in Elem.~F1, NVA, and Pres-Q, indicating that correctness, search behavior, and rendered quality provide complementary training signals.

\begin{figure}[h]
    \centering
    \includegraphics[width=0.85\textwidth]{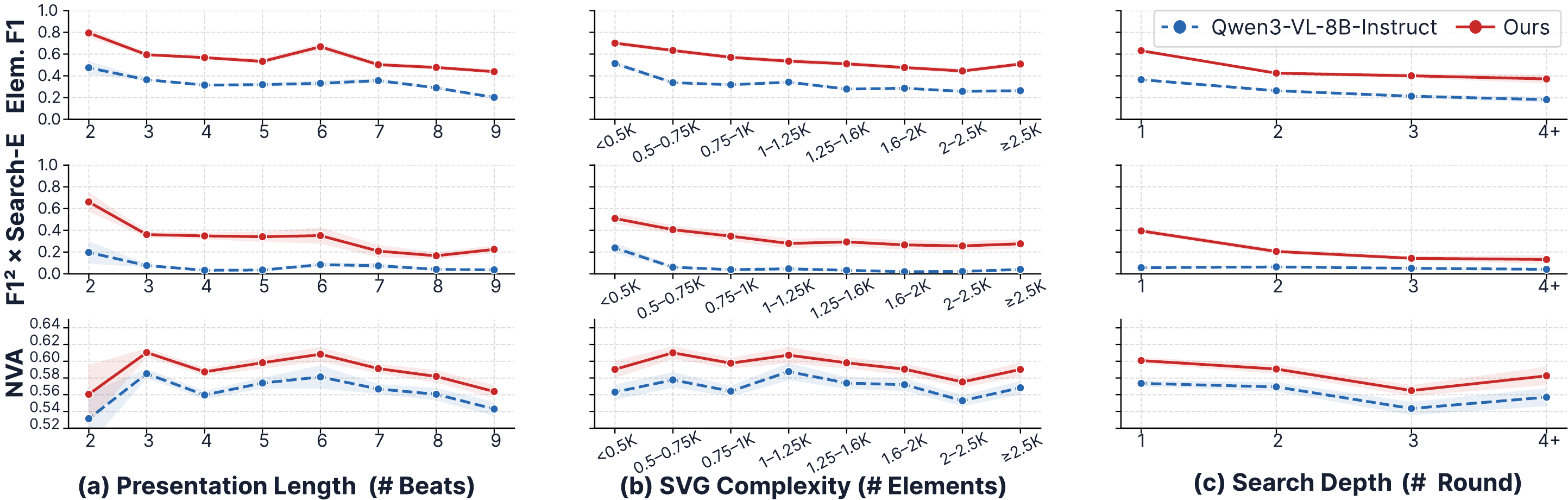}
    \caption{\textbf{FigAct-8B remains robust as task complexity increases.} FigAct-8B consistently outperforms the base model across presentation length, SVG complexity, and search depth.}
    \label{fig:scaling_ablation}
    \vspace{-1em}
\end{figure}

\textbf{FigAct Scales Gracefully with Increasing Task Complexity.}
As shown in Fig.~\ref{fig:scaling_ablation}, increasing presentation length, SVG
complexity, and search depth generally leads to performance degradation across Elem.~F1, Search-E, and NVA. Here, Search-E is weighted by grounding F1, so that efficient search also depends on whether it reaches the correct visual evidence. Despite the increasing difficulty, the FigAct-8B model
consistently outperforms the base model across all three scaling dimensions and generally exhibits milder degradation.
% The gap is particularly clear in Search-E, where FigAct maintains a substantial margin as presentations become longer, SVGs grow larger, and grounding requires deeper search.
This suggests that training improves not only grounding accuracy but also the ability to reach relevant evidence without excessive exploration as the search problem scales.

% ============================================================
\subsection{Human Evaluation}
\label{sec:human_eval}
\begin{wraptable}{r}{0.38\columnwidth}
\centering
\vspace{-8pt}
\scriptsize
\setlength{\tabcolsep}{1.8pt}
\renewcommand{\arraystretch}{1.05}

\resizebox{\linewidth}{!}{
\begin{tabular}{lcccccccc}
    \toprule

    % ---------- Comprehension Accuracy ----------
    \multicolumn{9}{c}{\textit{Comprehension Accuracy}} \\[-1pt]

    \multicolumn{9}{c}{
    \begin{tabular}{@{}cccc@{}}
        \makebox[1.15cm][c]{Text} &
        \makebox[1.15cm][c]{Highlight} &
        \makebox[1.15cm][c]{Prop.} &
        \makebox[1.15cm][c]{Ours} \\
        \makebox[1.15cm][c]{66.7\%} &
        \makebox[1.15cm][c]{75.6\%} &
        \makebox[1.15cm][c]{\textbf{86.7\%}} &
        \makebox[1.15cm][c]{83.3\%}
    \end{tabular}
    } \\

    \midrule

    % ---------- Likert header ----------
    \multirow{2}{*}{\textbf{Method}}
    & \multicolumn{7}{c}{\textbf{Likert Rating}}
    & \multirow{2}{*}{$\boldsymbol{\mu \pm \sigma}$} \\
    \cmidrule(lr){2-8}
    & 1 & 2 & 3 & 4 & 5 & 6 & 7 & \\
    \midrule

    % ---------- Comprehension Ease ----------
    \multicolumn{9}{c}{\textit{Comprehension Ease}} \\[-1pt]

    Text
    & \hB{6} & \hC{11} & \hD{17} & \hE{22}
    & \hD{17} & \hC{11} & \hB{6}
    & $4.00 \pm 1.61$ \\

    Highlight
    & \hA{3} & \hB{6} & \hC{11} & \hD{19}
    & \hE{22} & \hD{18} & \hC{11}
    & $4.66 \pm 1.56$ \\

    Prop.
    & \hA{1} & \hA{2} & \hB{7} & \hC{12}
    & \hD{20} & \hF{26} & \hE{22}
    & $\mathbf{5.38 \pm 1.40}$ \\

    Ours
    & \hA{1} & \hA{4} & \hB{8} & \hC{14}
    & \hE{23} & \hE{23} & \hD{17}
    & $5.12 \pm 1.45$ \\

    \midrule

    % ---------- Visual Clarity ----------
    \multicolumn{9}{c}{\textit{Visual Clarity}} \\[-1pt]

    Text
    & \hB{8} & \hC{13} & \hD{18} & \hE{22}
    & \hC{15} & \hB{9} & \hA{5}
    & $3.78 \pm 1.62$ \\

    Highlight
    & \hA{2} & \hA{5} & \hB{9} & \hD{17}
    & \hE{24} & \hD{20} & \hC{13}
    & $4.87 \pm 1.50$ \\

    Prop.
    & \hzero{0} & \hA{2} & \hA{5} & \hB{10}
    & \hD{19} & \hF{28} & \hF{26}
    & $\mathbf{5.60 \pm 1.29}$ \\

    Ours
    & \hzero{0} & \hA{2} & \hA{5} & \hC{11}
    & \hD{20} & \hF{29} & \hE{23}
    & $5.53 \pm 1.27$ \\

    \midrule

    % ---------- Visual Appeal ----------
    \multicolumn{9}{c}{\textit{Visual Appeal}} \\[-1pt]

    Text
    & \hB{10} & \hC{14} & \hD{18} & \hE{21}
    & \hC{14} & \hB{8} & \hA{5}
    & $3.66 \pm 1.66$ \\

    Highlight
    & \hA{2} & \hA{5} & \hC{11} & \hD{18}
    & \hE{23} & \hD{19} & \hC{12}
    & $4.78 \pm 1.50$ \\

    Prop.
    & \hzero{0} & \hA{2} & \hA{5} & \hB{9}
    & \hD{18} & \hF{29} & \hF{27}
    & $\mathbf{5.64 \pm 1.28}$ \\

    Ours
    & \hzero{0} & \hA{3} & \hB{6} & \hC{11}
    & \hE{21} & \hF{27} & \hE{22}
    & $5.43 \pm 1.34$ \\

    \bottomrule
\end{tabular}
}

\caption{Human evaluation results.}
\label{tab:human_eval}
\vspace{-8pt}

\end{wraptable}

\textbf{Protocol.}
We conduct a controlled user study with 18 graduate students from diverse disciplines. We compare four conditions: \textbf{Text}, which provides the textual explanation; \textbf{Highlight}, which additionally highlights the grounded evidence; \textbf{Proprietary MLLMs}, which uses Claude Fable $5.1$ to generate the full FigAct presentation, and \textbf{FigAct-8B (Ours)}. Text vs.\ Highlight measures the benefit of visual grounding, while Highlight vs.\ Ours isolates the effect of presentation actions. Each participant evaluates 20 question--figure pairs. We report \textit{comprehension accuracy} and 7-point Likert ratings for \textit{comprehension ease}, \textit{visual clarity}, and \textit{visual appeal}.
% Using a counterbalanced within-subject design, 

\textbf{Results.}
As shown in Table~\ref{tab:human_eval}, simply highlighting the grounded evidence already improves comprehension over text-only explanations ($75.6\%$ vs.\ $66.7\%$). The full FigAct presentation provides a further gain, reaching $83.3\%$ accuracy, with the clearest difference appearing in visual clarity ($5.53$ vs.\ $4.87$). This suggests that grounding helps users locate the relevant evidence, while presentation actions make that evidence easier to follow. Our model also remains close to the Proprietary baseline across the subjective measures.

% Additional analyses that can move to the supplementary material:
% - focal vs. context grounding F1
% - precision / recall separately
% - BeatSucc at multiple thresholds (0.5 / 0.75 / 0.9)
% - per-question-type breakdown
% - detailed grounding validity error taxonomy
% - flat vs. hierarchy absolute scores for each frozen MLLM
% - context-fit subset vs. full real-world set
% - judge rubric and human agreement
% - additional qualitative and failure cases

% \section{Limitations}

% FigAct currently generates question-conditioned presentations from the visual content of the scientific figure itself, without using the surrounding document context. As a result, questions that depend on the main text or broader scientific background can lead to less reliable responses. Since the three stages of FigAct can be used independently, one solution is to incorporate additional context at the planning stage. At the representation level, FigAct takes advantage of scientific figures with individually addressable visual elements. Grounding still relies on the rendered visual content, but the element structure provides a defined set of visual units to search and select and allows the grounded evidence to be directly manipulated by subsequent actions. Such structure is naturally provided by SVGs and vector-based PDFs, while raster-only figures require an additional vectorization or element decomposition step before being processed by FigAct.

\section{Conclusion}
We introduced FigAct, a framework that turns static scientific figures into question-conditioned visual presentations through a Plan-Ground-Act pipeline. Experiments on real-world scientific figures show that FigAct improves overall performance across both proprietary and open-weight MLLMs while substantially reducing SVG token usage. We further train FigAct-8B on a large-scale synthetic dataset and introduce a stage-aware GRPO objective to improve grounding correctness, search efficiency, and rendering quality. Human evaluation shows that linking explanations directly to visual evidence and presentation actions makes scientific figures easier to follow. Overall, FigAct demonstrates that scientific figures can serve not only as inputs for multimodal understanding, but also as active canvases for communicating model-generated explanations.

\section*{AI Use Statement}
We used generative AI tools to assist with writing and language polishing, code development, and debugging. All AI-assisted outputs were reviewed and verified by the authors, who take full responsibility for the content and results presented in this paper.

% We introduced FigAct, a framework that turns static scientific figures into question-conditioned visual presentations. FigAct structures presentation generation into the Plan-Ground-Act stage. Experiments on real-world scientific figures show that our proposed framework FigAct improves overall performance across both proprietary and open-weight MLLMs, while substantially reducing SVG token usage.
% We also trained a model, FigAct-8B, based on a large-scale synthetic dataset. 
% We propose a stage-aware GRPO to optimize grounding correctness, search efficiency, and rendering quality.
% Human evaluation further shows that directly connecting explanations with visual evidence and presentation actions can make scientific figures easier to follow. These results suggest that scientific figures can serve not only as inputs for multimodal understanding, but also as active canvases for communicating model-generated explanations.

% ============================================================
% References
% ============================================================

\bibliographystyle{iclr2027_conference}
\bibliography{iclr2027_conference}

% ============================================================
% Appendix
% ============================================================

\clearpage
\appendix

\section*{Appendix}

\startcontents[appendix]
\printcontents[appendix]{}{1}{}

\clearpage

\section{Additional Method Details}
\label{supp: method_detail}

\subsection{Geometry-Based SVG Forest Construction}

Real-world SVGs often contain highly fragmented primitives and unreliable DOM hierarchies, making direct grounding over the raw SVG both inefficient and structurally uninformative. We therefore reconstruct each SVG into a geometry-based forest that provides a compact hierarchical search space while preserving all original primitives.

\textbf{Text primitive grouping.}
SVG converters often represent a single text string as multiple glyph-level paths or \texttt{<use>} elements. We recover string-level units by matching text primitives to OCR regions. Each OCR bounding box is first mapped to the SVG coordinate system, and a text primitive is assigned to the OCR region with the largest bounding-box coverage when the coverage exceeds $0.30$. Primitives assigned to the same OCR region are merged into one text group, whose spatial extent is given by the union of their bounding boxes. OCR may miss small or stylized text. We therefore retain unmatched text primitives and cluster them using text-only spatial proximity. Let $c^{(i)}=(c_x^{(i)},c_y^{(i)})$ denote the center of the bounding box of primitive $i$. We define an anisotropic neighborhood with
\[
\varepsilon_x = 0.032W,
\qquad
\varepsilon_y = 0.028H,
\]
where $W$ and $H$ are the canvas width and height. Unmatched text primitives are connected when
\[
\left(
\frac{c_x^{(i)}-c_x^{(j)}}{\varepsilon_x}
\right)^2
+
\left(
\frac{c_y^{(i)}-c_y^{(j)}}{\varepsilon_y}
\right)^2
\leq 1.
\]
We extract connected components with Union--Find and treat each component as one fallback text group. Importantly, this proximity rule is applied only among text primitives and is never used to merge text with visual elements.

\begin{figure}[h]
    \centering
    \includegraphics[width=\textwidth]{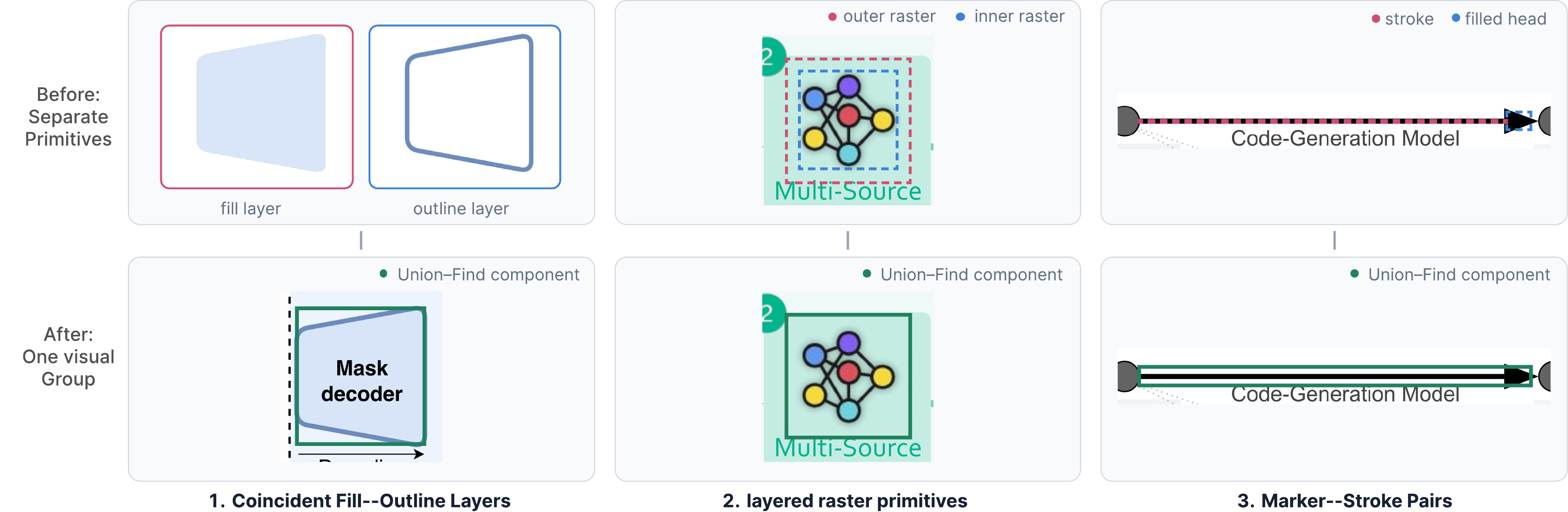}
\caption{
\textbf{Three rules for visual primitive grouping.}
Each example starts from separate SVG primitives and applies one of three high-confidence grouping rules: coincident fill--outline layers, layered raster renderings, or marker--stroke assembly. Primitives satisfying the corresponding rule are connected through Union--Find and consolidated into a single visual group.
}
    \label{fig:visual_grouping_rules}
\end{figure}

\textbf{Visual primitive grouping.}
For visual primitives, general spatial proximity is too permissive: nearby shapes, repeated colors, or touching connectors may belong to different semantic objects. We therefore introduce Union--Find edges only for three high-confidence same-object relations, illustrated in Fig.~\ref{fig:visual_grouping_rules}.
First, \emph{coincident fill--outline layers} are merged when two primitives have at least $0.90$ mutual bounding-box coverage and at least one primitive has a fill. Second, \emph{layered raster primitives}, such as nested masks or filtered copies of the same image, are merged when one covers at least $0.85$ of the other and their bounding-box area ratio is at least $0.50$. Third, \emph{marker--stroke pairs}, such as an arrowhead and its shaft, are merged when their paint attributes match and the marker is sufficiently close to the stroke. Each marker is attached only to its nearest eligible stroke, preventing one marker from connecting multiple neighboring strokes. 

\textbf{Containment-based hierarchy.}
After text and visual grouping, we organize all groups into a shared geometric forest. For a candidate child group $g_i$ and parent group $g_j$, we define
\[
C(g_i,g_j)
=
\frac{\operatorname{area}(B_i\cap B_j)}
     {\operatorname{area}(B_i)},
\]
where $B_i$ and $B_j$ are their bounding boxes. A group $g_j$ is considered a valid parent when $C(g_i,g_j)\geq0.85$. Among all valid candidates, we select the smallest enclosing group as the parent of $g_i$.
We further restrict visual parenthood to avoid spurious containment caused by large but mostly empty bounding boxes. In particular, an open stroke cannot serve as a parent solely because its bounding box contains other elements; a visual parent must contain a closed path or raster primitive that supports the enclosed region. The resulting parent relations form a deterministic, cycle-free forest while preserving the original SVG drawing order within each group.

\begin{figure}[t]
    \centering
    \includegraphics[width=0.92\textwidth]{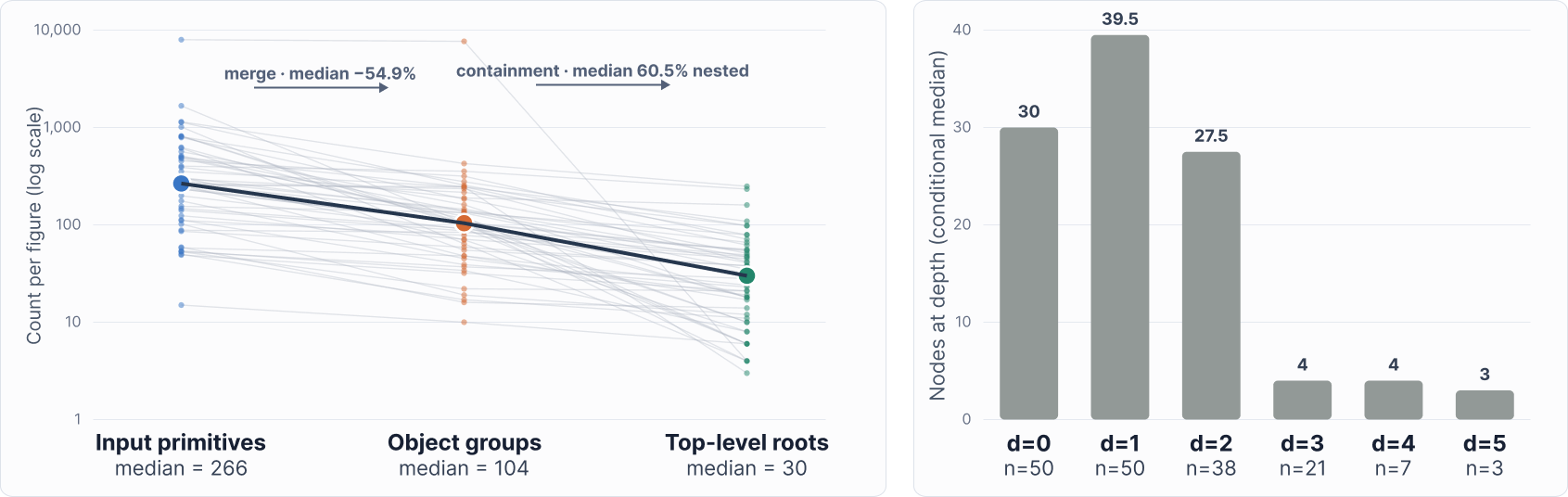}
\caption{
\textbf{SVG grouping statistics for 50 sampled figures.}
 Left: primitive, group, and top-level root counts
  for each figure. Right: median node counts at each containment depth among figures reaching that depth.
}
    \label{fig:group_statistics}
\end{figure}

\textbf{Grouping statistics.}
To quantify the effect of forest construction, we uniformly sample 50 figures without replacement from the real-world evaluation set and report the number of input SVG primitives, object-level groups after text and visual consolidation, and top-level roots after containment-based hierarchy construction. As shown in Figure~\ref{fig:group_statistics}, the median counts are 266 primitives, 104 groups, and 30 top-level roots per figure. When reductions are computed separately for each figure and then summarized by their median, grouping reduces the primitive count by $54.9\%$. Containment places $60.5\%$ of the groups below the root level,
resulting in $84.3\%$ fewer top-level entry points relative to the original primitive count. Importantly, containment does not discard these nested groups; it organizes them under a smaller set of roots. The resulting forests remain shallow, with a median maximum depth of 2 and an observed maximum depth of 5. These results show that the construction substantially reduces the flat SVG search space while preserving the grouped elements in a compact hierarchy for subsequent grounding.

\subsection{Action Space and Rendering}
\label{supp_action}

\begin{table*}[t]
\centering
\small
\setlength{\tabcolsep}{4pt}
\renewcommand{\arraystretch}{1.28}

\newcommand{\actionfig}[1]{%
    \includegraphics[width=2.20cm,keepaspectratio]{figs/appendix/actions/#1}%
}

\begin{tabular}{
    >{\raggedright\arraybackslash}m{1.80cm}
    >{\ttfamily\raggedright\arraybackslash}m{1.75cm}
    >{\ttfamily\raggedright\arraybackslash}m{4.00cm}
    >{\raggedright\arraybackslash}m{3.00cm}
    >{\centering\arraybackslash}m{2.25cm}
}
\toprule
\textbf{Type}
& \textbf{Action}
& \textbf{Parameters}
& \textbf{Usage}
& \textbf{Illustration} \\
\midrule

\textbf{View Control}
& camera
& target=<ids|page>
& Frame focal evidence with necessary context.
& \actionfig{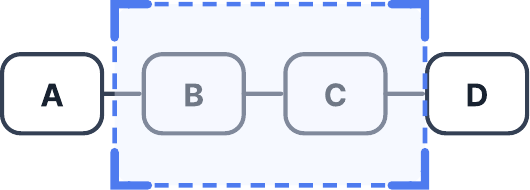}
\\

\midrule

\multirow{3}{2.15cm}{\textbf{Attention Control}}
& tint
& target=<ids> \newline
  [color=<color>]
& Assign a stable color to grounded evidence.
& \actionfig{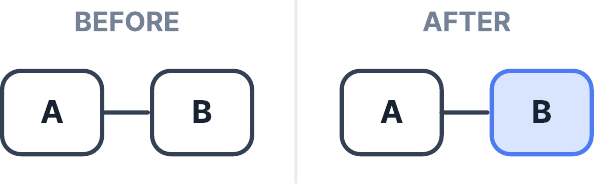}
\\
\cmidrule(lr){2-5}

& deemphasize
& target=<ids|page> \newline
  [keep=<ids>]
& Recede competing content while preserving context.
& \actionfig{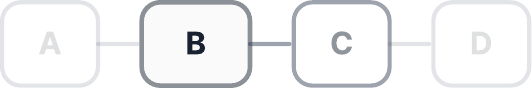}
\\
\cmidrule(lr){2-5}

& reemphasize
& target=<ids>
& Restore previously deemphasized content.
& \actionfig{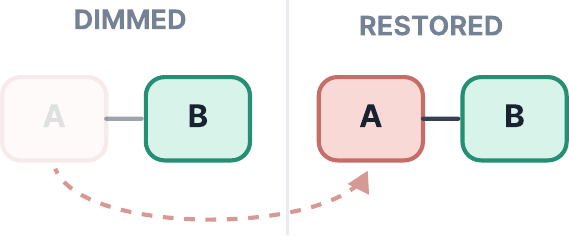}
\\

\midrule

\multirow{2}{2.15cm}{\textbf{Relational Guidance}}
& arrow
& connector=<connector-id>
& Highlight an existing relation or direction.
& \actionfig{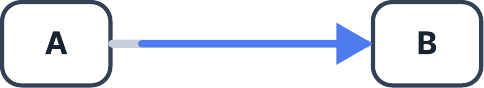}
\\
\cmidrule(lr){2-5}

& infer\_relation
& from=<id>, to=<id> \newline
  relation=<text>
& Add a labeled relation between grounded elements.
& \actionfig{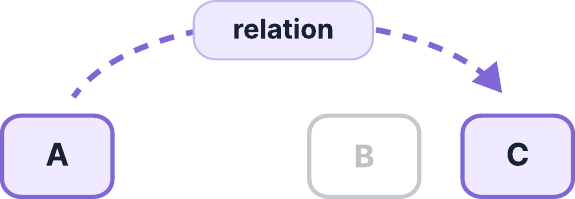}
\\

\midrule

\multirow{2}{2.15cm}{\textbf{Information Augmentation}}
& annotation
& target=<ids> \newline
  text=<text> \newline
  [placement=<auto|top|right| \newline
  bottom|left>]
& Attach a concise explanatory callout.
& \actionfig{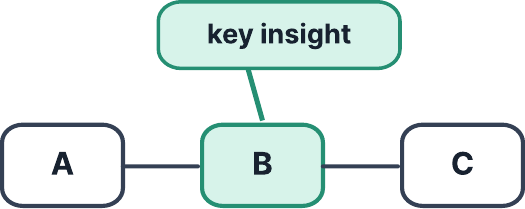}
\\
\cmidrule(lr){2-5}

& abstraction
& sources=<ids> \newline
  result.id=<id> \newline
  result.label=<text>
& Summarize grounded elements as a higher-level concept.
& \actionfig{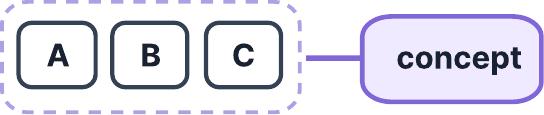}
\\

\bottomrule
\end{tabular}

\caption{
\textbf{Action space of FigAct.}
Actions are grouped by function types. For each action, we list its parameters, intended usage, and a representative rendered example. Square brackets denote optional parameters.
}
\label{tab:action_space}
\end{table*}

\textbf{Executable Action Space.}
The \textsc{Act} stage produces several executable visual actions to the grounded elements, used to guide readers' attention and augment the answer on the figure.
Our action space includes \texttt{tint}, \texttt{deemphasize}, \texttt{reemphasize}, \texttt{camera}, \texttt{arrow}, \texttt{infer\_relation}, \texttt{annotation}, and \texttt{abstraction}, as summarized in Table~\ref{tab:action_space}.
Actions within each beat are ordered, allowing multiple visual operations to form a presentation sequence.

\textbf{Semantic-to-SVG Rendering.}
The MLLM predicts visual actions at the semantic level, allowing it to focus on deciding which visual effect best communicates the current narration rather than generating error-prone, low-level SVG operations.
We implement a deterministic renderer that translates these semantic actions into element-specific SVG operations.
The same action can therefore be realized differently depending on the underlying visual element.
For example, \texttt{tint} modifies the fill of text and filled vector shapes, but operates on the stroke for connectors and open paths; for raster images, it creates an exterior highlight from the image mask.

\textbf{Stateful Action Execution.}
Rendering each beat independently would repeatedly reset the figure and reapply overlapping effects, resulting in unnecessary visual transitions and breaking the continuity of the presentation.
We therefore execute the action program statefully, treating the rendered state of one beat as the starting point of the next.
Concretely, the renderer maintains the active visual state of the figure, such as camera position, element visibility, tinting, and explanatory overlays.
When transitioning to a new beat, effects that remain relevant are preserved, effects that are no longer active are reverted to their original SVG state, and only newly introduced actions are executed.
For example, if two consecutive beats discuss different components within the same highlighted region, the camera and region-level emphasis can remain unchanged while only the component-specific highlight is updated.
This differential execution avoids redundant visual resets and produces smooth transitions across the presentation while keeping the original SVG state recoverable.

\section{Dataset Construction}
\label{sec:dataset}

\subsection{Synthetic Training Data}
\label{sec:synthetic_data}

\textbf{Source Figure Filtering.}
We construct our synthetic training data from VFIG-Data-Complex-Diagrams~\cite{he2026vfig}, which contains 60,429 scientific figures represented as SVGs.
Since these SVGs are reconstructed from raster figures, the raw collection contains conversion failures and low-quality examples.
We apply a two-stage filtering process.
We first remove invalid or degenerate SVGs using structural checks, including malformed canvases, insufficient visual content, extreme aspect ratios, unsupported SVG components, and external image references.
We then use Qwen3-VL-32B~\cite{qwen3technicalreport} to assess the rendered figures for visual quality, considering rendering completeness, legibility, layout, semantic coherence, and information content.
After filtering, we retain 12,509 figures for subsequent annotation.

\noindent
\begin{minipage}[t]{0.57\linewidth}
\vspace{0pt}

\textbf{Grounding Trajectory Reconstruction.}
The semantic SVG annotations in VFIG and its relatively compact SVG representations allow us to directly provide the full SVG to Qwen3-VL-32B for generating presentation plans, fine-grained grounding targets, and corresponding actions. These annotations specify \emph{what} visual evidence should ultimately be selected, while FigAct performs grounding through a multi-round coarse-to-fine search over the reconstructed SVG forest. Rather than separately annotating each intermediate search step, we deterministically reconstruct the trajectory from the final grounding targets.

Let $\mathcal{T}_b$ denote the final grounding targets for beat $b$, which we assume to be mutually non-ancestral (i.e., form an antichain in the forest). For each target $t\in\mathcal{T}_b$, the SVG forest defines a unique root-to-target path
\[
P(t)=(v_0,v_1,\ldots,v_L=t).
\]
We project each target onto its ancestors along this path. A candidate is assigned \texttt{Keep} if it is a final target, \texttt{Expand} if its subtree (i.e., its proper descendants $\mathrm{Desc}(v)$) contains a target, and \texttt{Delete} otherwise. Shared ancestors across multiple targets are merged via the set union over expanded children.
This yields a deterministic coarse-to-fine trajectory that progressively refines relevant branches until the final targets are reached, matching the grounding protocol used during inference.

\end{minipage}
\hfill
\begin{minipage}[t]{0.39\linewidth}
\vspace{0pt}
\footnotesize

\textbf{Algorithm 1: Grounding Trajectory Reconstruction}

\vspace{4pt}

\begin{algorithm}[H]
\caption{Grounding Trajectory Reconstruction}
\label{alg:traj}
\begin{algorithmic}[1]
\Require SVG forest $\mathcal{F}$, targets $\mathcal{T}_b$ (an antichain)
\Ensure trajectory $\tau_b$
\State $\mathcal{C} \gets \mathrm{Roots}(\mathcal{F})$
\State $\tau_b \gets [\,]$
\While{$\mathcal{C} \neq \emptyset$}
    \State $\mathcal{R} \gets \{\,\}$
    \State $\mathcal{C}' \gets \emptyset$
    \For{each $v \in \mathcal{C}$}
        \If{$v \in \mathcal{T}_b$}
            \State $\mathcal{R}[v] \gets \texttt{Keep}$
        \ElsIf{$\mathrm{Desc}(v)\cap\mathcal{T}_b \neq \emptyset$}
            \State $\mathcal{R}[v] \gets \texttt{Expand}$
            \State $\mathcal{C}' \gets \mathcal{C}' \cup \mathrm{Children}(v)$
        \Else
            \State $\mathcal{R}[v] \gets \texttt{Delete}$
        \EndIf
    \EndFor
    \State $\tau_b.\mathrm{append}(\mathcal{R})$
    \State $\mathcal{C} \gets \mathcal{C}'$
\EndWhile
\State \Return $\tau_b$
\end{algorithmic}
\end{algorithm}

\end{minipage}

\textbf{Structured Training Format.}
We provide a simplified training example below to illustrate the structured outputs of the \textsc{Plan}, \textsc{Ground}, and \textsc{Act} stages.
\begin{tcolorbox}[
breakable,
colback=gray!4,
colframe=gray!55,
colbacktitle=gray!70,
coltitle=white,
title={A Simplified Training Sample},
fonttitle=\bfseries,
arc=2mm,
boxrule=0.8pt
]
\footnotesize

\begin{verbatim}
{
  "question": "How are the encoder and decoder connected?",
  "plan": [
    {
      "beat": 1,
      "narration": "The encoder produces representations
                    that are passed to the decoder."
    }
  ],
  "ground": [
    {
      "beat": 1,
      "targets": ["elem_17", "elem_24"],
      "trajectory": [
        {"operation": "Expand", "group": "g_3"},
        {"operation": "Keep",   "group": "g_7"},
        {"operation": "Delete", "group": "g_8"}
      ]
    }
  ],
  "act": [
    {
      "beat": 1,
      "actions": [
        {"type": "tint", "target": "elem_17"},
        {"type": "camera", "target": "elem_24"}
      ]
    }
  ]
}
\end{verbatim}

\end{tcolorbox}

\subsection{Real-World Evaluation Data}
\label{sec:realworld_data}

\textbf{Figure Collection and SVG Conversion.}
To evaluate FigAct on real-world scientific figures, we collect figures from 2025-2026 papers published at NeurIPS, CVPR, MICCAI, CoRL, CHI, and UIST.
We consider papers with arXiv source packages available under the CC BY 4.0 license and download their complete \LaTeX{} sources.
From each source package, we identify candidate method figures using figure-related keywords such as \emph{architecture}, \emph{overview}, and \emph{method}.
We retain figures provided in PDF format, which preserves the vector structure needed for conversion into structured SVG representations.
We convert the collected PDFs to SVG using Inkscape with its Poppler-based PDF importer.
We choose this conversion pipeline because it provides a good balance between visual fidelity and structural cleanliness for \LaTeX-generated scientific figures.
In particular, it reliably preserves mathematical symbols and special characters by converting text into vector paths, avoiding font dependencies and glyph substitution during subsequent rendering.
At the same time, the converted SVGs retain the main vector geometry without introducing excessive editor-specific structures, making them suitable for downstream parsing and element-level operations.
This process results in 1,666 real-world SVG figures.

\begin{figure}[h]
    \centering
    \includegraphics[width=\textwidth]{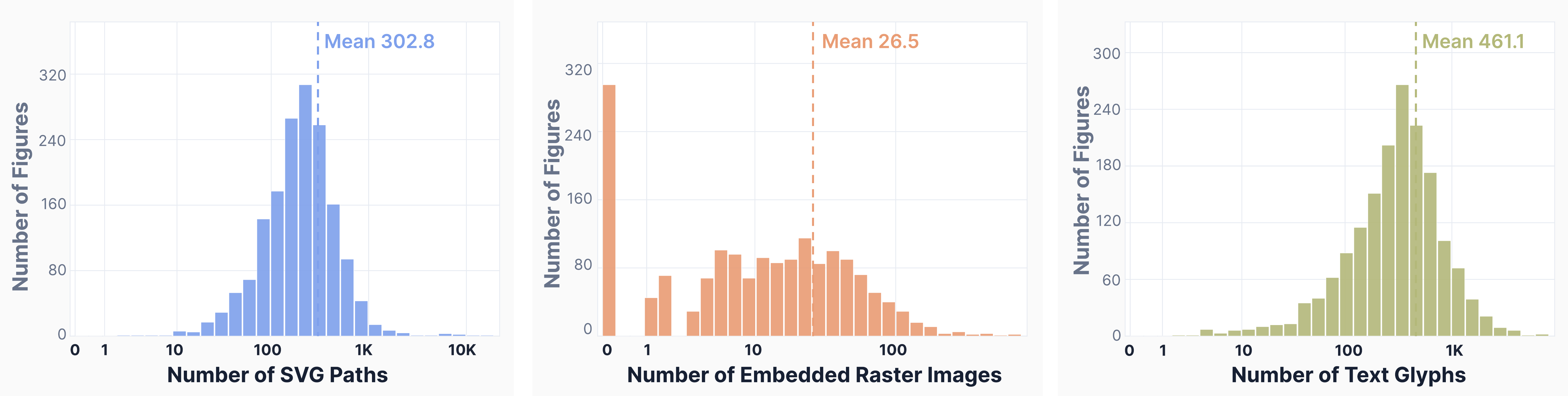}
    \caption{
\textbf{Structural complexity of 1,666 real-world scientific figures}, measured by
the numbers of SVG paths, embedded raster images, and text glyphs. Horizontal axes use a log-binned visualization scale.
}
    \label{fig:supp_real_svg_statistics}
\end{figure}

\paragraph{Real-world SVG Complexity.}
Real-world scientific figures contain substantially more
fine-grained and heterogeneous SVG content.
As shown in Fig.~\ref{fig:supp_real_svg_statistics}, a figure contains on average 302.8 SVG paths, 26.5 embedded raster images, and 461.1 text glyphs, with all three distributions exhibiting long tails. In particular, text is often represented as individual glyph paths and complex visual components may be decomposed into many low-level SVG
primitives. Consequently, directly exposing the complete SVG to an MLLM produces a large and poorly structured grounding space. These characteristics motivate our hierarchical SVG representation and coarse-to-fine grounding procedure, which allow FigAct to operate on meaningful groups without processing all low-level SVG elements at once.

\textbf{Model-assisted human annotation.}
Unlike the relatively compact SVGs in our synthetic data, real-world SVGs are substantially more complex, making it unreliable to directly provide the full SVG to an MLLM and use its output as ground-truth annotation. At the same time, annotating a complete presentation from scratch is costly: a single example can take up to 1.5 hours, as the annotator must repeatedly inspect selected SVG elements, verify the rendered effects, and organize the temporal order of actions across presentation beats.
We therefore adopt a model-assisted human annotation process. We first run FigAct with strong MLLMs, including Gemini 3.5 Flash and GPT-5.6, to generate candidate Plan--Ground--Act annotations and their rendered presentations. A human annotator then reviews these candidates, reconciles useful outputs across models, and iteratively corrects the plan, grounding selections, action choices, and temporal ordering while checking the rendered result after each revision. This process substantially reduces annotation effort while keeping the final supervision human-verified. Using this protocol, we construct 400 real-world presentation examples through one week of annotation.
% Supplementary fragment; no custom macros or table packages required.
% Evidence checked on 2026-09-19:
% training_scripts/train_vfig_qwen3vl8b_lora_8gpu.sh
% training_data/jsonl_data/training_jsonl_data/split_manifest.json
% /sensei-fs-3/users/sxiao/output/Qwen3-VL-8B-Instruct/
%   vfig-sft27-20260917_131707/v0-20260917-131734/args.json
%   .../checkpoint-6096/{adapter_config.json,trainer_state.json}
% reward/experiment/reward_ablation_d1d6_grpo/
%   {run_train.sh,dataset_manifest.json,stage_routing_patch.py,
%    ground_search_rewards.py,render_reward_worker.py,README.md}
% RL settings also checked against full/seed42-20260919-182237/
%   v0-20260919-182326/args.json under the reward-ablation-d1d6 output root.
% IMPORTANT: the loaded adapter has rank 16. The generic RL args.json
% lora_rank default of 8 does not describe that existing adapter.
% Runs currently use seed 42 only. Do not describe these as three-seed results.
% The 500-step setting is a budget; final completion/time/evaluation and
% the checkpoint used in the results tables must be documented when available.

\section{Training Details}
\label{sec:training_details}

\textbf{Training Data Split.}
We partition the synthetic data at the figure level to prevent different questions or presentations derived from the same figure from appearing across training and evaluation splits.
Specifically, 3,391 figures (27\%) are assigned to supervised fine-tuning, 7,869 figures (63\%) to reinforcement learning (RL), and 1,249 figures (10\%) to evaluation.
The SFT split is further converted into stage-specific training records for Plan, Ground, and Act.

\textbf{Supervised Fine-Tuning.}
We initialize FigAct from Qwen3-VL-8B-Instruct and perform SFT on Plan, Ground, and Act trajectories.
We apply LoRA with rank 16 and scaling factor 32 to all linear layers of the language model, while freezing the vision encoder and multimodal aligner.
The maximum sequence length is 24,576 tokens.
We train for 3 epochs on 8 GPUs with bfloat16 precision and SDPA attention, using a per-device batch size of 1 and 4 gradient accumulation steps.
We use AdamW with a learning rate of $1\times10^{-4}$, weight decay $0.1$, cosine decay, and a warmup ratio of $0.03$.
The resulting checkpoint initializes both the policy and reference adapters for RL.

\textbf{Multi-Turn Reinforcement Learning.}
Starting from the SFT checkpoint, we apply stage-aware GRPO to Ground and Act while keeping the presentation Plan fixed.
Each rollout performs multi-turn hierarchical Ground search, then constructs a fresh Act context from the final predicted grounding and samples the executable action program.
The Ground context retains the complete search history, whereas Act does not inherit intermediate Ground turns.
Ground and Act are optimized separately, with their losses weighted by the number of generated loss-bearing tokens in each stage.
We use 20,876 presentations for RL training and reserve 3,317 figure-disjoint presentations for held-out evaluation.
For each prompt, we sample $G=8$ rollouts with temperature $0.8$, top-$p=0.95$, and top-$k=50$.
The context length is capped at 65,536 tokens, with up to 24,576 generated tokens per turn subject to the remaining context budget.
We use a learning rate of $1\times10^{-6}$, cosine decay, warmup ratio $0.03$, and KL coefficient $\beta=0.04$.
Training uses bfloat16 and DeepSpeed ZeRO-3 on 8 GPUs, with 4 GPUs for policy optimization and colocated rollout generation and 4 GPUs for rendering and reward computation. The entire training process completes in approximately 35 hours.

\textbf{Rendering-Based Reward Computation.}
For each rollout, we validate the generated Ground and Act outputs and render each presentation beat with headless Chromium through Playwright at $896\times896$ resolution and 8 FPS.
We uniformly sample 8 frames per beat and use PE-Core~\cite{bolya2026perception} to measure visual--text alignment with the fixed beat narration.
Visual and text embeddings are $\ell_2$-normalized, and their cosine similarities are aggregated to obtain the beat-level rendering reward.
The final rendering reward is averaged across beats.
Because narration is fixed during RL, this reward reflects changes in grounding and visual actions.

\textbf{Stage-Specific Reward Optimization.}
Ground tokens receive grounding correctness $r_{\mathrm{grd}}$, search efficiency $r_{\mathrm{search}}$, and rendering quality $r_{\mathrm{render}}$ with weights $1.0$, $0.5$, and $0.5$, respectively.
Act tokens receive only $r_{\mathrm{render}}$.
For each prompt, each reward channel is independently normalized across the $G=8$ rollouts to obtain its group-relative advantage.
The normalized advantages are then weighted and routed to their corresponding stages, preventing reward channels with different scales from being coupled through normalization of a single aggregated reward.

\section{Evaluation Protocol}

\subsection{Evaluation Metrics}
\label{app:metrics}

\subsubsection{Planning Evaluation}
We evaluate the presentation plan $P = \{n_t\}_{t=1}^{T}$ along two complementary dimensions: factuality (precision) and information coverage (recall). Both metrics produce normalized scores bounded in $[0, 1]$, where higher values denote superior quality.

\textbf{Factuality.}
Following prior work on factual verification in multimodal settings~\cite{xu-etal-2023-critical}, we decompose each generated narration unit $n_t$ into atomic factual claims $\mathcal{C}(n_t)$ and verify whether each claim is supported by the figure $F$:
\[
\mathrm{Factuality}(P)
=
\frac{
\sum_{t=1}^{T}
\sum_{c \in \mathcal{C}(n_t)}
\mathbf{1}[F \models c]
}{
\sum_{t=1}^{T} |\mathcal{C}(n_t)|
},
\]
where $\mathbf{1}[\cdot]$ indicates whether claim $c$ is entailed by figure $F$ ($F \models c$).

\textbf{Coverage.}
Inspired by question-answering evaluation protocols like QuestEval~\cite{scialom-etal-2021-questeval}, we assess how thoroughly plan $P$ addresses user query $q$ given figure $F$. We construct a set of target probe questions and reference answers $\mathcal{Q}(F, q) = \{(q_i, a_i)\}_{i=1}^{K}$ covering key figure details. An evaluator answers each probe using only the concatenated narration text $N$ from plan $P$, yielding predicted answer $\hat{a}_i$. Coverage is defined as:
\[
\mathrm{Coverage}(N, F, q)
=
\frac{1}{K}
\sum_{i=1}^{K}
\mathbf{1}[\hat{a}_i \equiv a_i],
\]

\subsubsection{Visual Grounding Evaluation}
\label{supp:eval_grounding}
\textbf{Completeness (Comp.).}
Completeness measures beat-level presence—evaluating whether the MLLM successfully outputs predictions for all expected presentation beats in the reference plan. Given $T_{\mathrm{ref}}$ as the set of ground-truth beat IDs and $T_{\mathrm{pred}}$ as the set of predicted beat IDs for a sample, Beat Completeness is defined as:
\[
\mathrm{Comp} = \frac{|T_{\mathrm{ref}} \cap T_{\mathrm{pred}}|}{|T_{\mathrm{ref}}|}.
\]

\textbf{Element F1 (Elem. F1).}
Element F1 evaluates element-level precision and recall within each beat against human-curated ground-truth SVG element sets. For the $t$-th presentation beat, let $\hat{G}_t$ and $G_t^*$ denote the predicted and reference terminal element sets, respectively. We compute element-level precision $\mathsf{Prec}_t$ and recall $\mathsf{Rec}_t$ as:
\[
\mathsf{Prec}_t = \frac{|\hat{G}_t \cap G_t^*|}{|\hat{G}_t|}, \qquad
\mathsf{Rec}_t = \frac{|\hat{G}_t \cap G_t^*|}{|G_t^*|}, \qquad
\mathrm{F1}_t = \frac{2 \cdot \mathsf{Prec}_t \cdot \mathsf{Rec}_t}{\mathsf{Prec}_t + \mathsf{Rec}_t}.
\]
Beat-level F1 scores are macro-averaged across beats to yield the overall sample score.

\textbf{Search Efficiency (Search-E).}
Search Efficiency assesses whether the model reaches the target evidence without incurring excessive search cost across the SVG hierarchy. We define the search cost $C$ as the total number of unique child nodes exposed during hierarchical search. For beat $t$, let $\hat{C}_t$ and $C_t^*$ denote the predicted and oracle reference search costs, respectively. We evaluate search efficiency as the ratio of minimal required cost to actual cost:
\[
\mathrm{Search\text{-}E}_t = \min\left(1, \frac{C_t^* + 1}{\hat{C}_t + 1}\right).
\]

% \textbf{Grounding Validity.}
% Let $\mathcal{E}(F)$ denote all valid SVG element IDs in figure $F$. We define
% \[
% V_t=
% \frac{
% |\hat G_t\cap\mathcal{E}(F)|
% }{
% \max(1,|\hat G_t|)
% },
% \qquad
% \mathrm{GroundValid}
% =
% \frac{1}{T}\sum_{t=1}^{T}V_t.
% \]
% This metric captures hallucinated or nonexistent SVG references.

\subsubsection{Acting Evaluation}
\label{supp:eval_act}

\textbf{Execution Rate (Exec.).}
Execution Rate measures the executability and validity of the generated action script $A$. To be executable, an action script must pass: 1) \textbf{Schema \& Constraint Check:} Action types, required parameters, and relational constraints must be valid. 2) \textbf{Grounding ID Resolution:} All targeted visual element IDs must exist within the resolved grounding element set or previously instantiated runtime objects.
Execution Rate is defined as the fraction of samples in the benchmark that yield successfully rendered, error-free videos:
\[
\mathrm{Exec} = \frac{N_{\mathrm{rendered}}}{N_{\mathrm{total}}}.
\]
Scripts that violate structural or rendering constraints are flagged as unexecutable and excluded from the subsequent NVA evaluation.

\textbf{Narration--Video Alignment (NVA).}
For each beat, we measure the cross-modal similarity between its narration $n_t$ and rendered video segment $v_t$ using a frozen video--text encoder:
\[
\mathrm{NVA}
=
\frac{1}{T}
\sum_{t=1}^{T}
\cos\!\left(
\phi_{\mathrm{text}}(n_t),
\phi_{\mathrm{video}}(v_t)
\right).
\]

\subsubsection{End-to-End Presentation Quality}
\textbf{End-to-End Presentation Quality (Pres-Q).}
We render each generated SVG animation into video and employ a
rubric-based MLLM judge to evaluate the final presentation from three complementary dimensions:
\emph{Content Quality}, which measures whether the visual presentation
emphasizes the figure evidence supporting the narration;
\emph{Visual Effectiveness}, which measures whether the applied visual actions
make the intended information clear and easy to follow; and
\emph{Temporal Coherence}, which measures whether the visual changes form a consistent and coherent progression throughout the presentation.
We render the generated SVG animations as H.264 videos at
$988\times520$ resolution and 8 FPS, with each 3-second clip containing 24 frames. We use Qwen3-VL-32B as the video judge and provide it with the rendered presentation together with the corresponding narration. The judge independently assigns a score in $[0,1]$ to each dimension, and we report their mean as the end-to-end presentation quality.

\subsection{Human Evaluation Protocol}
\label{sec:human_protocol}

\begin{figure}[t]
    \centering
    \includegraphics[width=\linewidth]{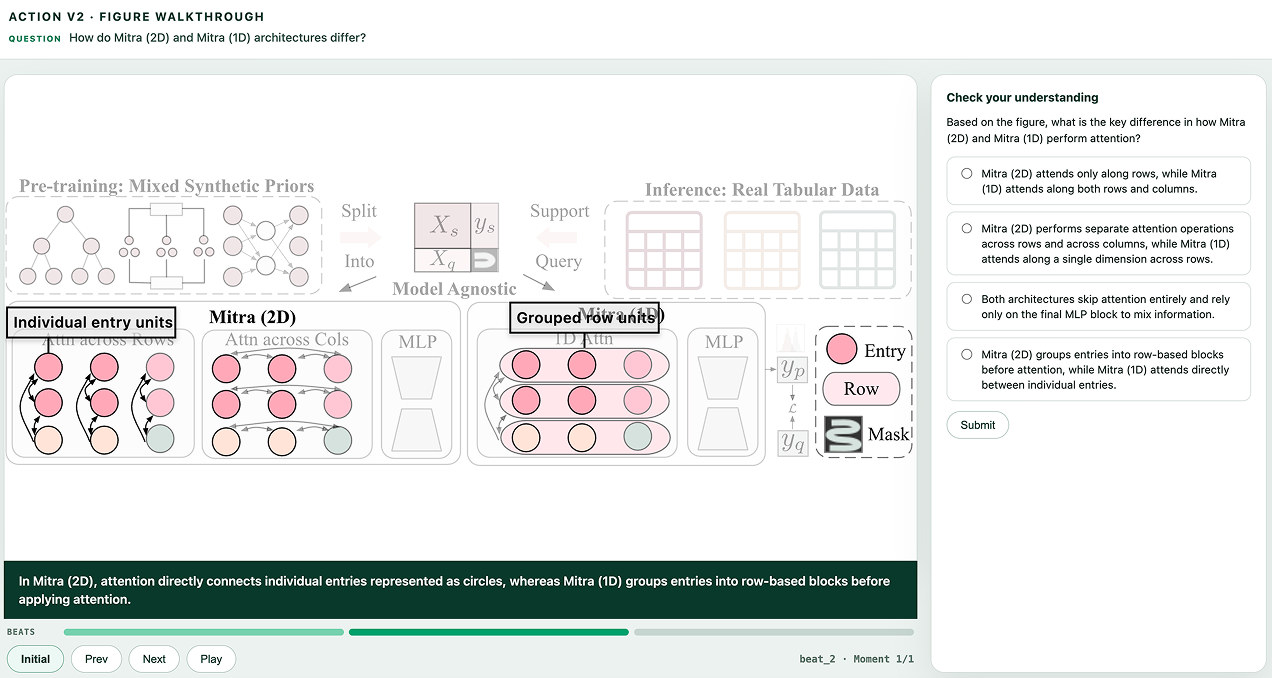}
    \caption{
    \textbf{Interface used in the human evaluation.} Participants view a question-conditioned explanation of a scientific figure and then answer a comprehension question.
    }
    \label{fig:human_eval_interface}
\end{figure}

\textbf{Study Procedure.}
Each evaluation trial consists of two stages, as illustrated in
Fig.~\ref{fig:human_eval_interface}.
Participants first view the assigned explanation for a question--figure pair
and then answer a comprehension question designed to test whether they
understood the information conveyed by the explanation.
They subsequently rate their viewing experience using the three subjective
measures described below.
Question--figure pairs are counterbalanced across presentation conditions such
that a participant does not evaluate multiple presentations of the same pair,
avoiding direct repetition of the same content.
The order of evaluation trials is randomized for each participant.

\textbf{Metric Definitions.}
\emph{Comprehension Accuracy} is an objective measure of whether the participant correctly answers the comprehension question associated with each trial, and is reported as the percentage of correct responses.
The remaining three measures capture the aspects of the viewing experience.
\emph{Comprehension Ease} asks how easy it was to understand the information needed to answer the question.
\emph{Visual Clarity} asks how clearly the presentation guided attention to the relevant visual evidence in the figure.
\emph{Visual Appeal} asks how visually appealing the resulting presentation was.
All three subjective measures use a 7-point Likert scale, with higher values indicating more positive ratings.
% rollout ablation

% qualitative exmaples

% validation syntehtic 的实验? 

\section{Additional Experimental Results}
\label{supp_experiment}

\subsection{Ablation Studies}

\paragraph{Effect of Context and Isolated Views.}
We study the visual representation used during hierarchical grounding on two representative MLLMs: the proprietary Gemini 3.5 Flash and the open-weight Qwen3-VL-8B.
For both models, we apply FigAct directly at inference time, allowing us to isolate the effect of the rendering views themselves.
Each candidate SVG group is presented through a \emph{context view}, which preserves its location and surrounding structure in the full figure, an \emph{isolated view}, which removes surrounding elements, or both views together.

As shown in Table~\ref{tab:dual_view_ablation}, isolated rendering consistently improves grounding performance over the context-only view for both models, with especially large gains on the stricter GS@$0.75$ metric, which measures the proportion of presentation beats with an element-level F1 score of at least \(0.75\).
This indicates that explicitly separating candidate elements from surrounding visual clutter is important for fine-grained SVG grounding.
Combining both views further improves grounding accuracy for both models, highlighting that their joint use is the most effective design for FigAct.

\begin{table}[h]
\centering
\footnotesize
\setlength{\tabcolsep}{2.5pt}
\renewcommand{\arraystretch}{1.0}

\begin{tabular}{llcccc}
\toprule
\textbf{Model} & \textbf{View}
& Elem. F1 $\uparrow$
& GS@0.5 $\uparrow$
& GS@0.75 $\uparrow$
& Search-E $\uparrow$ \\
\midrule

\multirow{3}{*}{Gemini 3.5 Flash}
& Context only
& 60.99 & 64.47 & 40.53 & 89.50 \\

& Isolated only
& \underline{67.09}
& \underline{86.87}
& \underline{58.63}
& \underline{95.18} \\

& \textbf{Dual view}
& \textbf{68.85}
& \textbf{89.63}
& \textbf{59.82}
& \textbf{95.35} \\

\midrule

\multirow{3}{*}{Qwen3-VL-8B}
& Context only
& 17.75 & 21.87 & 3.90 & 63.75 \\

& Isolated only
& \underline{26.42}
& \underline{25.10}
& \underline{16.17}
& \textbf{69.73} \\

& \textbf{Dual view}
& \textbf{28.35}
& \textbf{27.83}
& \textbf{16.92}
& \underline{68.26} \\

\bottomrule
\end{tabular}

\caption{
Ablation of context and isolated rendering for hierarchical grounding.
Both models use FigAct only at inference time without FigAct-specific training.
Best results are shown in bold and second-best results are underlined.
}
\label{tab:dual_view_ablation}
\end{table}

\paragraph{GRPO vs. Stage-Aware GRPO.}
\begin{wraptable}{r}{0.45\columnwidth}
    \centering
    \vspace{-8pt}
    \scriptsize
    \setlength{\tabcolsep}{2.6pt}
    \renewcommand{\arraystretch}{1.05}

    \resizebox{\linewidth}{!}{
    \begin{tabular}{lcccc}
        \toprule
        \textbf{Setting}
        & Elem. F1 $\uparrow$
        & Search-E $\uparrow$
        & NVA $\uparrow$
        & Pres-Q $\uparrow$ \\
        \midrule

        SFT only
            & 52.67
            & 83.04
            & 69.68
            & 80.56 \\

        GRPO
            & \underline{61.41}
            & \underline{84.62}
            & \underline{70.83}
            & \underline{81.53} \\

        \textbf{Stage-Aware GRPO}
            & \textbf{62.27}
            & \textbf{84.68}
            & \textbf{72.25}
            & \textbf{82.16} \\

        \bottomrule
    \end{tabular}
    }

    \caption{
        Comparison between standard GRPO and our stage-aware reward optimization.
    }
    \label{tab:grpo_ablation}
    \vspace{-8pt}
\end{wraptable}
We compare standard GRPO with our stage-aware reward optimization under the same SFT initialization and training configuration.
Standard GRPO first aggregates the reward components and performs group-relative normalization on the combined reward.
In contrast, our stage-aware variant normalizes each reward channel independently and routes the resulting advantages to the corresponding Ground and Act stages.
As shown in Table~\ref{tab:grpo_ablation}, both RL variants substantially improve over SFT, confirming that reward-based optimization provides additional gains beyond supervised training.
Stage-aware GRPO further improves Elem. F1 from 61.41 to 62.27, NVA from 70.83 to 72.25, and Pres-Q from 81.53 to 82.85, while achieving slightly lower search efficiency than standard GRPO.

\paragraph{Effect of Rollout Group Size.}
We study the effect of the rollout group size $G$ by varying the number of sampled rollouts per prompt while keeping the remaining training configuration fixed.
As shown in Table~\ref{tab:rollout_ablation}, increasing the group size from 4 to 8 yields consistent gains, while further increasing it to 16 leads to only marginal changes. Importantly, the effect is not monotonic across all metrics, with NVA peaking at \(G=8\), suggesting that larger rollout groups mainly improve the stability of relative optimization rather than uniformly improving every downstream objective.

\begin{table}[h]
\centering
\footnotesize
\setlength{\tabcolsep}{3.2pt}
\renewcommand{\arraystretch}{1.05}
\begin{tabular}{cccccc}
\toprule
$G$
& Elem. F1 $\uparrow$
& Search-E $\uparrow$
& Exec. $\uparrow$
& NVA $\uparrow$
& Pres-Q $\uparrow$ \\
\midrule

4
& 61.51
& 83.38
& 96.64
& 71.07
& 79.88 \\

8
& 62.27
& 84.68
& 97.10
& 72.25
& 82.16 \\

16
& 62.46
& 84.89
& 97.18
& 71.87
& 82.46 \\

\bottomrule
\end{tabular}

\caption{
Effect of rollout group size on FigAct performance.
}
\label{tab:rollout_ablation}
\end{table}

\paragraph{Generalization from Synthetic to Real-World Figures.}
We evaluate the Qwen3-VL-8B-based FigAct model on both a held-out synthetic set drawn from the training data construction pipeline and our real-world scientific figure benchmark.
Since the two sets differ substantially in figure complexity and data source, their absolute scores are not directly comparable; instead, we examine how the gains from SFT and RL transfer across the two settings.
As shown in Table~\ref{tab:synthetic_real}, SFT substantially improves grounding on both datasets.
RL provides further gains on the synthetic set and, more importantly, continues to improve grounding and action quality on real-world figures.
The consistent improvement across both settings suggests that the learned Ground and Act behaviors are not limited to the synthetic training distribution.
\begin{table}[t]
\centering
\footnotesize
\setlength{\tabcolsep}{3.5pt}
\renewcommand{\arraystretch}{1.05}

\begin{tabular}{llcccc}
\toprule
\textbf{Dataset} & \textbf{Model}
& Elem. F1 $\uparrow$
& Search-E $\uparrow$
& NVA $\uparrow$
& Pres-Q $\uparrow$ \\
\midrule

\multirow{3}{*}{Synthetic}
& Qwen3-VL-8B
& 28.75
& --
& 67.37
& \underline{78.91} \\

& + SFT
& \underline{61.68}
& \underline{85.38}
& \underline{67.84}
& 78.85 \\

& + SFT + RL
& \textbf{63.83}
& \textbf{85.92}
& \textbf{78.21}
& \textbf{79.85} \\

\midrule

\multirow{3}{*}{Real-world}
& Qwen3-VL-8B
& 9.85
& --
& 67.92
& 70.18 \\

& + SFT
& \underline{60.67}
& \underline{83.04}
& \underline{69.68}
& \textbf{82.47} \\

& + SFT + RL
& \textbf{62.27}
& \textbf{84.68}
& \textbf{72.25}
& \underline{82.16} \\

\bottomrule
\end{tabular}

\caption{
Evaluation on held-out synthetic and real-world scientific figures.
Synthetic and real-world scores are reported separately because the two datasets differ in distribution and difficulty.
}
\label{tab:synthetic_real}
\end{table}

\subsection{Additional Analysis}

\subsubsection{Action Distribution.}
We observe clear differences in how models use the FigAct action space.
Proprietary MLLMs exhibit distinct presentation styles. GPT favors annotation, Gemini more often uses deemphasizing and camera control, while Claude distributes its actions more evenly across several operations.
Open-weight models show a different pattern, with much stronger concentration on a single action. Qwen3-VL-8B assigns 94.0\% of its actions to tint, while InternVL3.5-8B assigns 73.3\% to the same operation.
This pattern changes substantially in FigAct-8B.
Tint usage drops from 94.0\% to 16.2\%, while the model makes much greater use of deemphasizing, annotation, camera control, and arrows.
The resulting action distribution is considerably less concentrated and more similar to the broader action usage observed in proprietary MLLMs, suggesting that FigAct-8B learns to use the available action space in a more varied way when constructing presentations.

\begin{table*}[t]
\centering
\footnotesize
\setlength{\tabcolsep}{3.2pt}
\renewcommand{\arraystretch}{1.05}

\begin{tabular}{lrrrrrrrr}
\toprule
\textbf{Model}
& \textbf{Deemph.}
& \textbf{Annot.}
& \textbf{Tint}
& \textbf{Camera}
& \textbf{Arrow}
& \textbf{Reemph.}
& \textbf{Infer Rel.}
& \textbf{Abstract.} \\
\midrule

Claude
& 20.7
& 32.5
& 6.2
& 4.9
& 9.0
& 13.3
& 6.6
& 6.9 \\

Gemini
& 39.4
& 21.0
& 7.4
& 14.1
& 7.4
& 9.2
& 1.0
& 0.5 \\

GPT
& 18.9
& 45.6
& 3.7
& 2.8
& 11.4
& 14.5
& 2.1
& 0.9 \\

InternVL3.5-8B
& 1.7
& 1.3
& 73.3
& 16.2
& 2.9
& 4.6
& 0.0
& 0.0 \\

Qwen3-VL-8B
& 4.3
& 0.0
& 94.0
& 0.8
& 0.5
& 0.4
& 0.0
& 0.0 \\

Qwen3.5-9B
& 25.1
& 9.8
& 58.5
& 1.5
& 1.1
& 3.2
& 0.8
& 0.0 \\

\midrule

\textbf{FigAct-8B (Ours)}
&30.5
& 31.9
& 16.2
& 8.5
& 9.1
& 1.2
& 2.0
& 0.6 \\

\bottomrule
\end{tabular}

\caption{
Distribution of visual actions generated by different models.
All values are percentages (\%) of the total generated actions for each model and sum to approximately 100\% due to rounding.
}
\label{tab:action_distribution}
\end{table*}

\subsubsection{Failure Cases and Limitations.}
We observe two representative failure modes of FigAct-8B on real-world scientific figures.

\begin{figure}[h]
    \centering
    \includegraphics[width=0.68\textwidth]{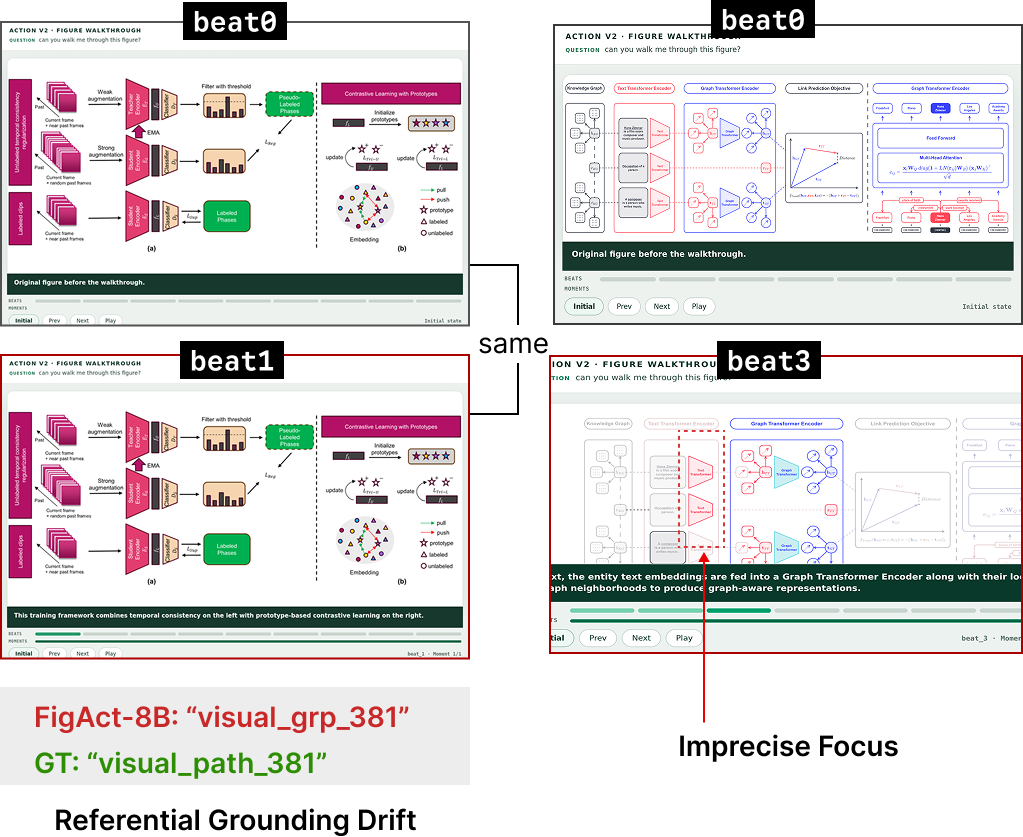}
\caption{
\textbf{Representative failure cases of FigAct-8B on real-world scientific figures.}
(a) \textbf{Identifier binding error.}
The model predicts a near-match but nonexistent SVG identifier, causing the
intended visual action to fail and leaving the next beat visually unchanged.
(b) \textbf{Imprecise focus.}
The model identifies the correct semantic region but selects a region that is
too broad, leaving distracting neighboring content in the focused view.
}
    \label{fig:failure_cases}
\end{figure}

The first is an \emph{identifier binding error}.
As shown in Figure~\ref{fig:failure_cases} left, the model predicts
\texttt{visual\_grp\_381}, while the valid target is
\texttt{visual\_path\_381}.
The prediction is very close to the correct identifier, but the referenced
element does not actually exist at runtime.
As a result, the renderer cannot apply the intended action, and the walkthrough
moves to the next beat with almost no visible change in the figure.
We see this as a real-world grounding issue rather than a failure to understand
the content itself: the model appears to identify the intended element, but
does not always bind it to the exact SVG symbol required for execution.
This is more likely to occur in real-world SVGs, where many identifiers differ
only by small changes in element type, suffix, or index, compared with the more
distinct identifier patterns in our synthetic data.

The second failure mode is \emph{imprecise focus}.
In Figure~\ref{fig:failure_cases} right, the model focuses on the correct part of the
figure, but the selected region is still too broad and retains distracting
neighboring content.
The issue is therefore not simply whether the model finds the right region,
but whether it selects the right level of visual detail for presentation.
These examples show that reliable figure walkthroughs require both exact
element binding and well-calibrated visual scope.

\textbf{Limitations}.
FigAct currently generates question-conditioned presentations from the visual content of the scientific figure itself, without using the surrounding document context. As a result, questions that depend on the main text or broader scientific background can lead to less reliable responses. Since the three stages of FigAct can be used independently, one solution is to incorporate additional context at the planning stage. At the representation level, FigAct takes advantage of scientific figures with individually addressable visual elements. Grounding still relies on the rendered visual content, but the element structure provides a defined set of visual units to search and select and allows the grounded evidence to be directly manipulated by subsequent actions. Such structure is naturally provided by SVGs and vector-based PDFs, while raster-only figures require an additional vectorization or element decomposition step before being processed by FigAct.

\section{Qualitative Results}
\label{supp: qualitative}

\begin{figure}[h]
    \centering
    \includegraphics[width=\textwidth]{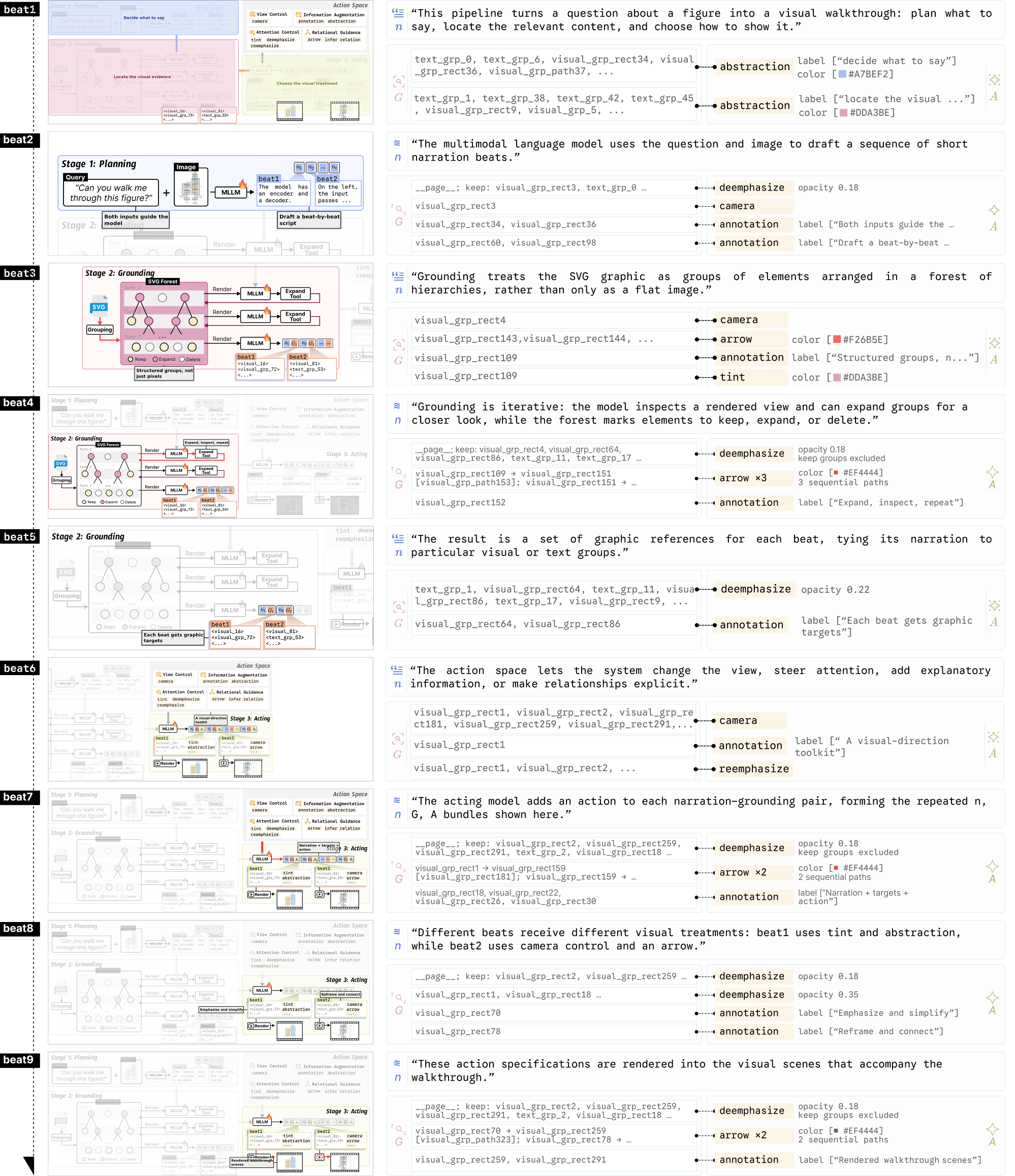}
    \caption{
\textbf{Complete FigAct presentation for the example introduced in Figure~\ref{fig:framework_overview}.} 
}
    \label{fig:supp_main_example}
\end{figure}

\begin{figure}[t]
    \centering
    \includegraphics[width=\textwidth]{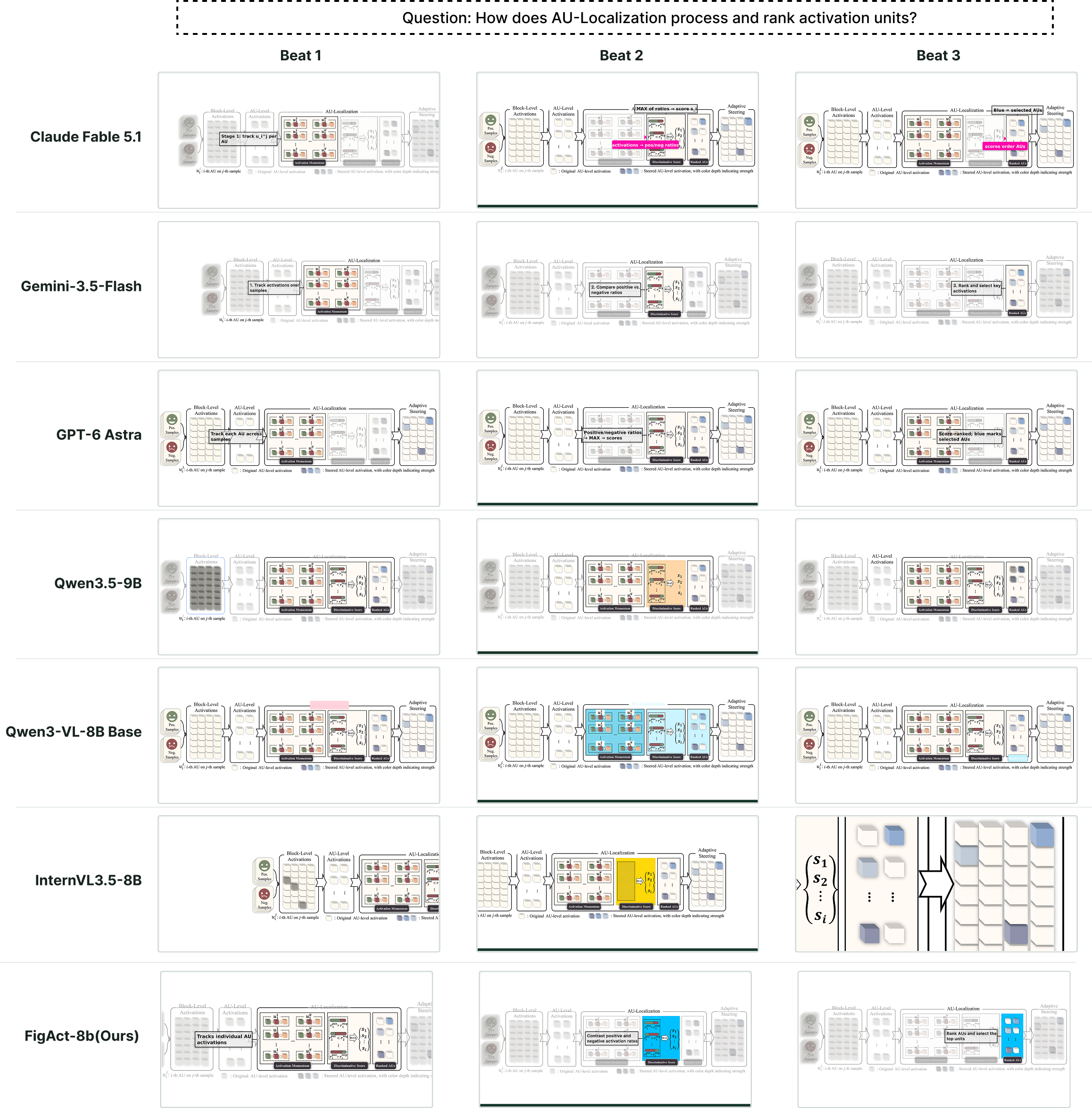}
\caption{
\textbf{Qualitative comparison of question-conditioned presentations across models.}
Rows show different models and columns show three presentation beats for the same figure and question.
The target narration is:
\textbf{Beat 1:} \texttt{AU-Localization starts with Activation Momentum, which tracks individual activations $u_i^j$ for the $i$-th AU on the $j$-th sample};
\textbf{Beat 2:} \texttt{Discriminative Score compares the positive and negative ratios, $r_i^{\mathrm{pos}}$ and $r_i^{\mathrm{neg}}$, and applies the \texttt{MAX} operator to obtain scores $s_1,\ldots,s_i$};
\textbf{Beat 3:} \texttt{these scores determine the ordering in Ranked AUs, with selected activations highlighted in blue.}
The scientific figure used as input is reproduced from~\cite{feng2026fine}.}
    \label{fig:supp_compare_example_1}
\end{figure}

\begin{figure}[t]
    \centering
    \includegraphics[width=\textwidth]{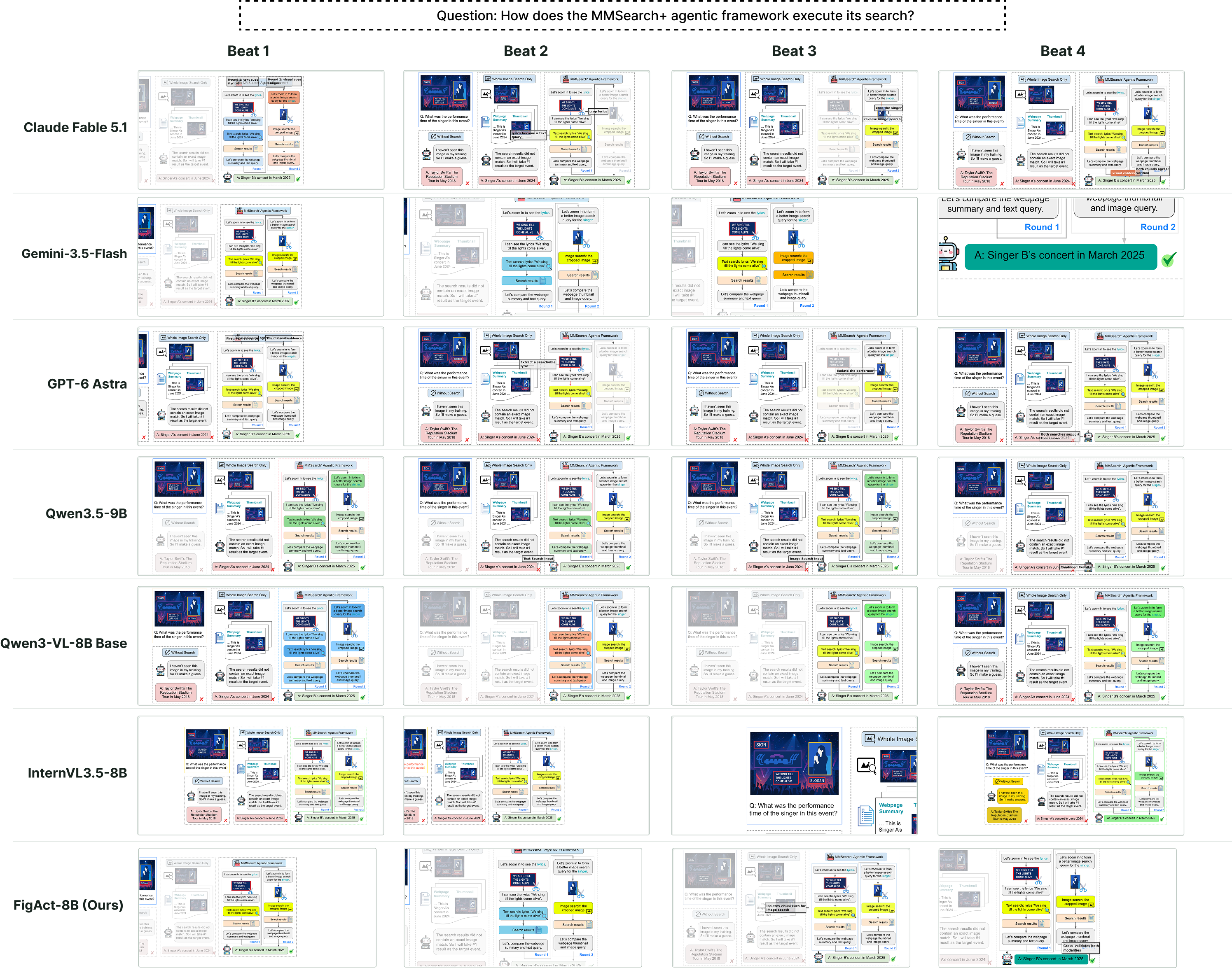}
\caption{
\textbf{Qualitative comparison of question-conditioned presentations across models.}
Rows show different models and columns show four presentation beats for the same figure and question.
The target narration is:
\textbf{Beat 1:} \texttt{The framework operates in two distinct search rounds: Round 1 focuses on text cues like lyrics, while Round 2 focuses on visual cues like the singer};
\textbf{Beat 2:} \texttt{In Round 1, the agent zooms in and crops the lyrics 'We sing till the lights come alive', then runs a text search to find matching song info};
\textbf{Beat 3:} \texttt{In Round 2, the agent crops the singer's silhouette to perform a reverse image search, gathering visual matches for the performer};
\textbf{Beat 4:} \texttt{Finally, the agent compares the results from both rounds to confirm the correct event and performance time.
}The scientific figure used as input is reproduced from~\cite{tao2026mmsearch}.
}
    \label{fig:supp_compare_example_2}
\end{figure}

\begin{figure}[t]
    \centering
    \includegraphics[width=\textwidth]{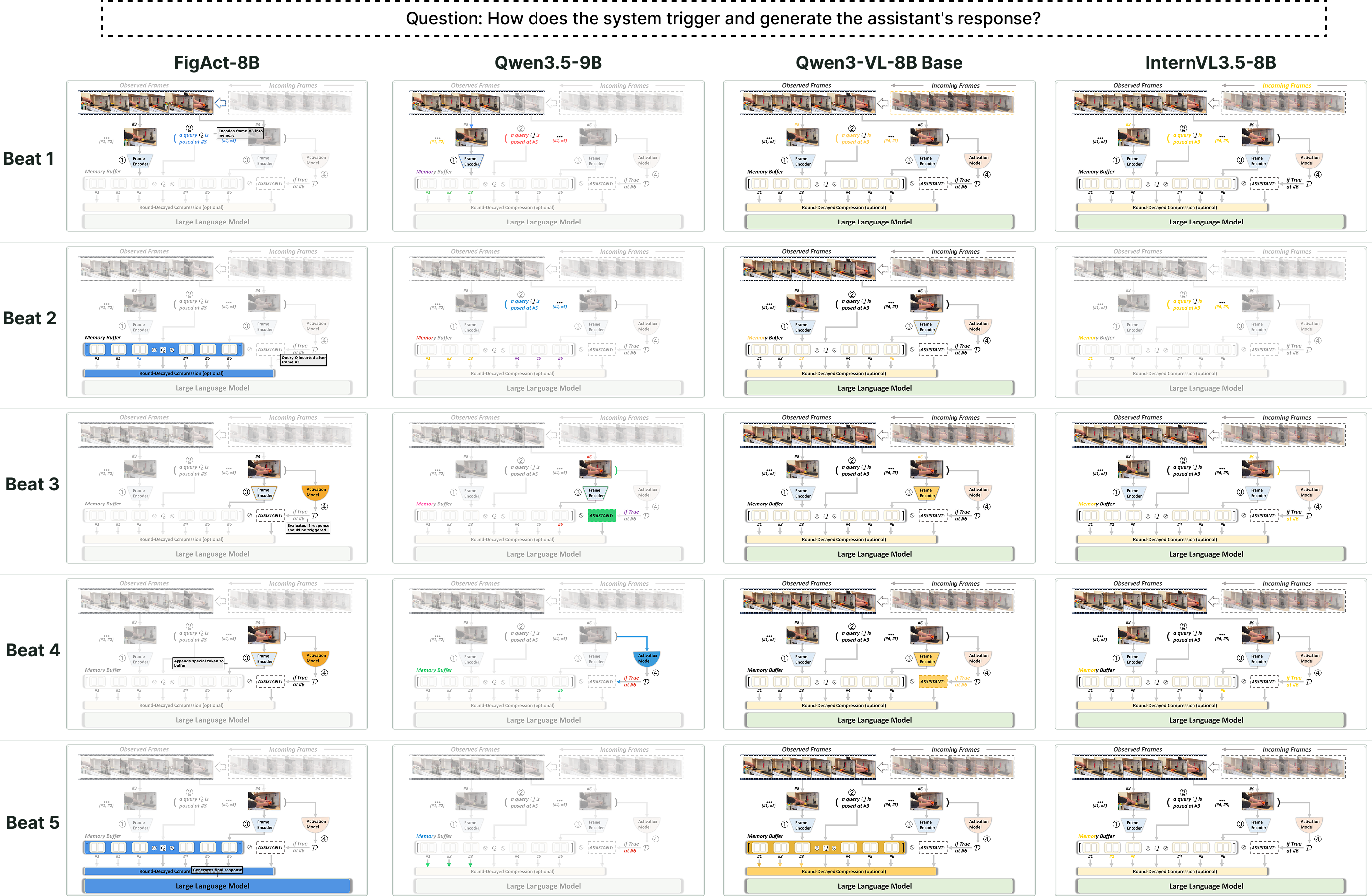}
    \caption{
        \textbf{Qualitative comparison of question-conditioned presentations across models.}
        The target narration is:
        \textbf{Beat 1:} \texttt{A query Q is posed at observed frame 3, which is processed by a frame encoder to create a representation in the memory buffer};
        \textbf{Beat 2:} \texttt{The memory buffer sequences representations of frames 1 through 6, with the query Q interleaved right after frame 3};
        \textbf{Beat 3:} \texttt{As frame 6 is observed, it is sent to both a frame encoder and an Activation Model to determine if a response should be triggered};
        \textbf{Beat 4:} \texttt{If the Activation Model outputs 'True' at frame 6, an 'ASSISTANT:' token is appended to the memory buffer};
                \textbf{Beat 5:} \texttt{The compiled buffer is optionally compressed via Round-Decayed Compression before being processed by the Large Language Model to generate the final response.} The scientific figure used as input is reproduced from~\cite{wang2026streambridge}.
    }
    \label{fig:supp_compare_example_3_1}
\end{figure}

\begin{figure}[t]
    \centering
    \includegraphics[width=\textwidth]{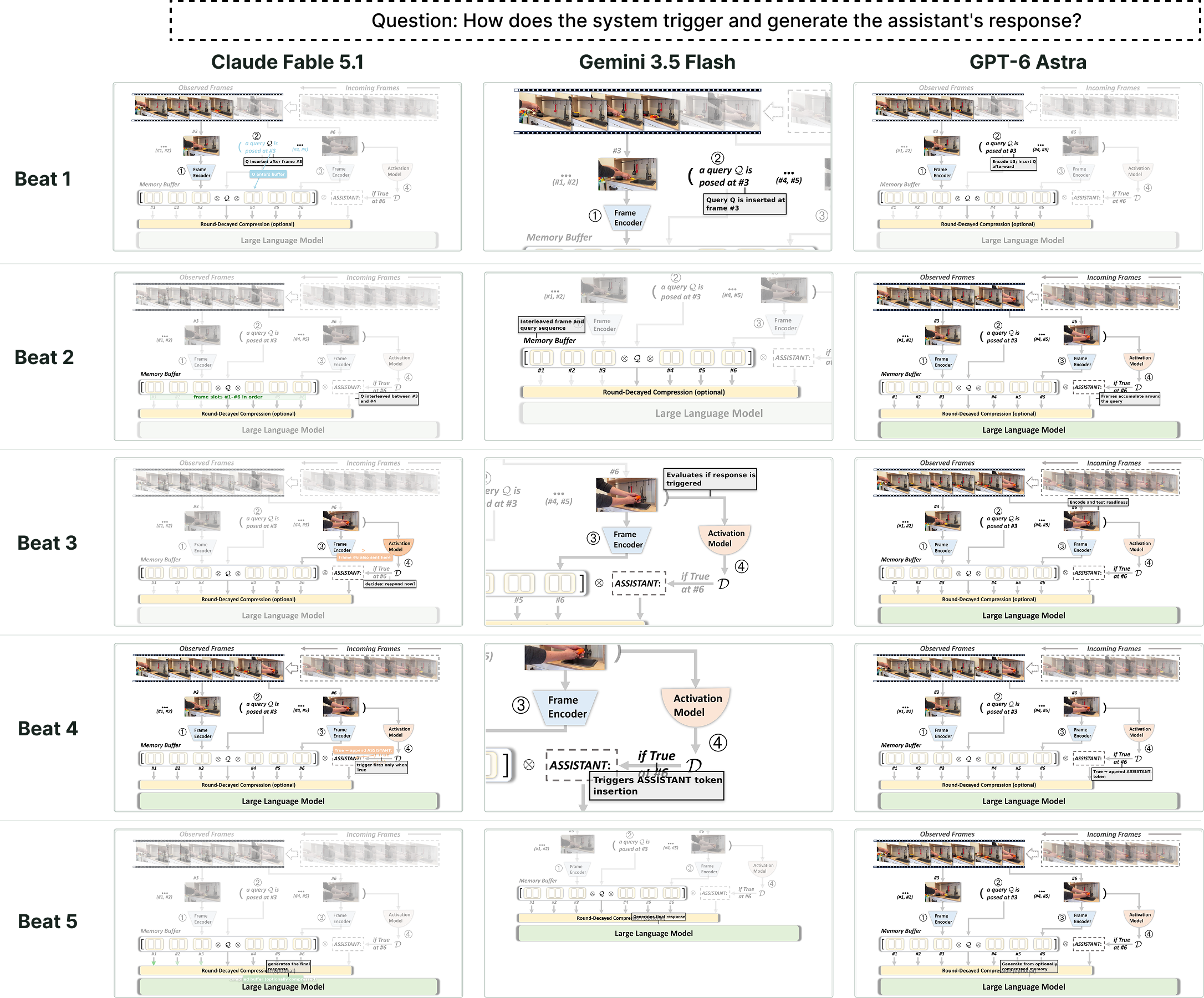}
    \caption{
        \textbf{Qualitative comparison of question-conditioned presentations across models.}
        The target narration is:
        \textbf{Beat 1:} \texttt{A query Q is posed at observed frame 3, which is processed by a frame encoder to create a representation in the memory buffer};
        \textbf{Beat 2:} \texttt{The memory buffer sequences representations of frames 1 through 6, with the query Q interleaved right after frame 3};
        \textbf{Beat 3:} \texttt{As frame 6 is observed, it is sent to both a frame encoder and an Activation Model to determine if a response should be triggered};
        \textbf{Beat 4:} \texttt{If the Activation Model outputs 'True' at frame 6, an 'ASSISTANT:' token is appended to the memory buffer};
                \textbf{Beat 5:} \texttt{The compiled buffer is optionally compressed via Round-Decayed Compression before being processed by the large language model to generate the final response. }The scientific figure used as input is reproduced from~\cite{wang2026streambridge}.
    }
    \label{fig:supp_compare_example_3_2}
\end{figure}

We show the complete example from Figure~\ref{fig:framework_overview} in Figure~\ref{fig:supp_main_example}, along with additional comparison examples in Figure~\ref{fig:supp_compare_example_1}, Figure~\ref{fig:supp_compare_example_2}, Figure~\ref{fig:supp_compare_example_3_1}, and Figure~\ref{fig:supp_compare_example_3_2}.

% Preamble: \input{supp_prompt/prompt_style.tex}
\clearpage
\section{Prompt Templates}
\label{app:prompt-templates}
We present the prompts for \textsc{Plan},
\textsc{Ground}, and \textsc{Act} as follows.
\begin{tcolorbox}[
breakable,
colback=figactplan,
colframe=figactplandark!65,
colbacktitle=figactplandark,
coltitle=white,
title={Prompt 1: Plan --- Atomic Walkthrough Planning},
fonttitle=\bfseries,
arc=2mm,
boxrule=0.8pt
]
\footnotesize

\textbf{Input.}
\texttt{<image>} (complete figure); user question: \texttt{<question>}.

You are designing a reader-facing visual walkthrough for one figure from a research paper.
Understand the figure globally, then help the reader build a mental model and progressively reach its important insights.

\textbf{Planning Rules}

\begin{enumerate}
\item Infer the figure's purpose, major regions, and dominant flow or comparison.
Organize by explanatory logic, not by scanning every object.
Distinguish visible evidence from interpretation; be cautious when labels or relationships are ambiguous.

\item Start with the main input/output, central contrast, or dominant pipeline.
Establish the overall picture before secondary branches, losses, feedback paths, or implementation details.
Resolve likely confusions, including repeated modules, crossing arrows, and training versus inference.

\item Build the sequence around insights supported by visible evidence, rather than an inventory of objects.

\item Make every beat atomic: one compact idea and one coherent visual focus, with no internal steps.
Split changes of region, multi-stage paths, and overview-to-detail transitions into separate beats.
A beat may focus on one object, a tightly related group, a direct comparison, or a connector and its endpoints.

\item Keep narration brief and speakable, normally one sentence.
Give the visual direction a concrete visible identity and location so it can later be grounded to SVG elements.
Use one dominant attention change per beat.
Establish objects before tracing their relationships.

\item Align purpose, narration, visual presentation, and takeaway around the same idea.
The visual direction is an attention instruction, not a caption;
the takeaway is an understanding, not another instruction.
Avoid redundant beats and low-value detail, while covering all major regions needed for the explanation.
\end{enumerate}

\textbf{Return only JSON:}

\begin{verbatim}
{"status":"plan", "question":"<question>", "beats":[{
  "beat_id":"beat_1",
  "purpose":"distinct explanatory job",
  "beat_narration":"brief spoken narration",
  "visual_presentation":"focused visual direction with identity and location",
  "reader_takeaway":"single understanding to retain"
}]}
\end{verbatim}

\end{tcolorbox}

\begin{tcolorbox}[
breakable,
colback=figactground,
colframe=figactgrounddark!65,
colbacktitle=figactgrounddark,
coltitle=white,
title={Prompt 2: Ground --- Top-Level Selection},
fonttitle=\bfseries,
arc=2mm,
boxrule=0.8pt
]
\footnotesize

\textbf{Input.}
Atomic walkthrough plan; top-level candidate manifest
(\texttt{page}, \texttt{svg\_id}, \texttt{n\_children});
\texttt{<image>} (clean full figure), followed by annotated candidate pages.

Ground every atomic beat to the supplied top-level SVG candidates.
Each annotated page shows complete-figure context with candidate boxes and exact SVG IDs above,
and the same candidates isolated as cards below.

Imagine presenting with these materials.
Select all independently annotated elements needed by each beat,
preserving their visual/presentation order.

\textbf{Selection Rules}

\begin{itemize}
\item \texttt{focal\_svg\_ids}: primary visible evidence directly discussed or animated.
\item \texttt{context\_svg\_ids}: only structure needed to preserve identity, containment,
relationship, orientation, or the integrity of a compound visual.
\item Imagine all unselected content hidden.
Add the minimum context needed to keep the focal evidence clear.
Context is not another list of topics.
Prefer one meaningful container when the complete container is necessary.
\end{itemize}

For each selected SVG, choose independently:

\begin{itemize}
\item \texttt{keep}: usable at this level.
\item \texttt{expand}: relevant but too coarse; inspect its direct children next.
\item \texttt{keep\_and\_expand}: retain the complete candidate as evidence or context while also inspecting its direct children.
\end{itemize}

Do not keep a parent merely because it contains the evidence.
If the visual direction requires internal modules, endpoints, arrows, labels,
or an internal sequence, expand the candidate when it has children.

Use only exact IDs from the manifest.
Identify beats by \texttt{beat\_id} and copy \texttt{beat\_narration} exactly.
Omit purpose, visual presentation, reader takeaway, storyboard, and steps.
Keep reasons on SVG items, not at beat level.

\textbf{Return only JSON:}

\begin{verbatim}
{"status":"ground_round1", "beats":[{
  "beat_id":"beat_1",
  "beat_narration":"copied exactly",
  "focal_svg_ids":[{
    "svg_id":"exact_id",
    "decision":"keep|expand",
    "role":"role in beat",
    "reason":"visible evidence"}],
  "context_svg_ids":[{
    "svg_id":"exact_id",
    "decision":"keep|expand",
    "context_type":"container|label|connector|endpoint|landmark|composite_part",
    "role":"context role",
    "reason":"what becomes unclear without it"}]
}]}
\end{verbatim}

\end{tcolorbox}

\begin{tcolorbox}[
breakable,
colback=figactground,
colframe=figactgrounddark!65,
colbacktitle=figactgrounddark,
coltitle=white,
title={Prompt 3: Ground --- Recursive Parent Expansion},
fonttitle=\bfseries,
arc=2mm,
boxrule=0.8pt
]
\footnotesize

\textbf{Input.}
Current question (when available); opened parent ID; requested beat/channel pairs;
direct-child manifest; \texttt{<image>} (clean opened-parent render),
followed by annotated direct-child pages.

Refine only the requested beats for the specified opened parent.
Select only that parent's direct children from the manifest.
Return one entry per requested \texttt{beat\_id},
copying \texttt{beat\_narration} exactly from its supplied beat object.

Each annotated page shows the opened-parent context with boxed children
and exact SVG IDs above, and isolated direct-child cards below.
Use context to locate each child and cards to inspect its content.
Select every independently annotated child needed by the beat,
preserving presentation order.

\textbf{Refinement Rules}

\begin{itemize}
\item Return both \texttt{focal\_svg\_ids} (primary evidence)
and \texttt{context\_svg\_ids} (support needed for identity, containment,
relationships, orientation, or visual integrity).

\item Repeat the context-completeness check at this finer level:
imagine all unselected children hidden and add only those whose absence
makes the focal result unclear or incomplete.

\item If the parent came from \texttt{context\_svg\_ids},
do not promote its children to focal evidence unless the beat's
\texttt{visual\_presentation} directly requires them.

\item Choose independently for each child:
\texttt{keep} if usable;
\texttt{expand} if still too coarse;
\texttt{keep\_and\_expand} if the complete child must remain visible
while its children are inspected.

\item Use exact direct-child IDs.
Do not add beats, storyboard, or steps.
Omit purpose, visual presentation, and reader takeaway.
Retain reasons only on individual SVG items.
\end{itemize}

\textbf{Return only JSON:}

\begin{verbatim}
{"status":"ground_expand", "parent_id":"unchanged", "beats":[{
  "beat_id":"beat_1",
  "beat_narration":"copied exactly",
  "focal_svg_ids":[{
    "svg_id":"exact_child_id",
    "decision":"keep|expand",
    "role":"primary role",
    "reason":"specific visible evidence"}],
  "context_svg_ids":[{
    "svg_id":"exact_child_id",
    "decision":"keep|expand",
    "context_type":"container|label|connector|endpoint|landmark|composite_part",
    "role":"context role",
    "reason":"what becomes unclear without it"}]
}]}
\end{verbatim}

\end{tcolorbox}

\begin{tcolorbox}[
breakable,
colback=figactact,
colframe=figactactdark!65,
colbacktitle=figactactdark,
coltitle=white,
title={Prompt 4: Act --- Executable Visual Presentation},
fonttitle=\bfseries,
arc=2mm,
boxrule=0.8pt
]
\footnotesize

\textbf{Input.}
Grounded walkthrough; \texttt{<image>} (clean full figure),
followed by annotated beat panels (one to three panels).

You are a visual walkthrough director.
Convert the grounded walkthrough into an executable visual presentation.
Inspect the complete Ground result and all images first.
Return every beat exactly once, in its original order.
Communicate each beat's message visually without requiring the viewer to read its narration.

\textbf{ID Rules}

Use only SVG IDs listed in \texttt{focal\_svg\_ids} or
\texttt{context\_svg\_ids} anywhere in the complete grounded walkthrough,
preferring the current beat's IDs.
Preserve IDs exactly; never invent or rewrite source IDs.

\textbf{Renderer-Native Actions}

\textbf{Attention}

\begin{itemize}
\item \texttt{tint}: use color only for a clear semantic distinction.
Prefer pastel colors, consistent semantic roles, and complete grounded groups.

\begin{verbatim}
{"action":"tint", "target":"id"|["id"],
 "style"?:{"color":"#RRGGBB"}}
\end{verbatim}

\item \texttt{deemphasize}: recede competing content; the renderer sets the strength.

\begin{verbatim}
{"action":"deemphasize",
 "target":"__page__"|"id"|["id"],
 "style"?:{"keep":["id"]}}
\end{verbatim}

\item \texttt{reemphasize}: restore deemphasized content without changing its color.

\begin{verbatim}
{"action":"reemphasize", "target":"id"|["id"]}
\end{verbatim}
\end{itemize}

\textbf{View}

\begin{itemize}
\item \texttt{camera}: reframe only when the current view impedes understanding.

\begin{verbatim}
{"action":"camera", "target":"id"|["id"]|"__page__"}
\end{verbatim}
\end{itemize}

\textbf{Relations}

\begin{itemize}
\item \texttt{arrow}: animate a connector already drawn in the figure.

\begin{verbatim}
{"action":"arrow",
 "from":"id"|["id"],
 "to":"id"|["id"],
 "connector":"existing_connector_id",
 "style"?:{"color":"#RRGGBB", "duration":0.9}}
\end{verbatim}

\item \texttt{infer\_relation}: add a short labeled relation only when necessary
and not already shown by an existing connector.

\begin{verbatim}
{"action":"infer_relation",
 "from":"id",
 "to":"id",
 "relation":"short relation",
 "style"?:{"color":"#RRGGBB", "duration":0.9}}
\end{verbatim}
\end{itemize}

\textbf{Augmentation}

\begin{itemize}
\item \texttt{annotation}: add an insight, preferably under six words,
rather than copying a visible label.

\begin{verbatim}
{"action":"annotation",
 "target":"id"|["id"],
 "text":"short insight",
 "placement"?:"auto"|"top"|"right"|"bottom"|"left",
 "style"?:{"color":"#RRGGBB","variant":"black_label"}}
\end{verbatim}

\item \texttt{abstraction}: group at least two independent elements
sharing a property or semantic role.

\begin{verbatim}
{"action":"abstraction",
 "sources":["id","id"],
 "result":{"id":"runtime_id","label":"short label"},
 "style"?:{"color":"#RRGGBB"}}
\end{verbatim}
\end{itemize}

\textbf{Presentation Rules}

Use the smallest clear action set, in execution order,
and maintain coherence across beats.
Preserve colors that encode information;
use other actions if recoloring changes meaning.
Use at most three camera actions in the walkthrough.
Combine \texttt{deemphasize}, \texttt{tint}, and \texttt{camera}
in one beat only if all three are essential.

Write \texttt{visual\_presentation} to describe the actions'
visible result and attention order.
Do not include SVG IDs or copy the upstream presentation instruction.

\textbf{Return only valid JSON:}

\begin{verbatim}
{"beats":[{
  "beat_id":"beat_1",
  "visual_presentation":"visible result and attention order",
  "actions":[]
}]}
\end{verbatim}

\end{tcolorbox}

\end{document}